\documentclass[5p,twocolumn,times,preprint,sort&compress]{elsarticle}

\usepackage{hyperref,lineno}
\modulolinenumbers[10]

\usepackage{setspace}
\usepackage{xspace}
\usepackage{adjustbox,fancyvrb,framed} 
\usepackage{enumitem} 		   
\usepackage{tabularx,booktabs} 
\usepackage{bm}				   
\usepackage{mathtools} 		   
\usepackage{lipsum}            
\usepackage{color,soul}        
\usepackage{amsmath}           
\usepackage{natbib}            
\usepackage{tikz}              
\usepackage{esvect}  
\usepackage{pifont}
\newcommand{\cmark}{\ding{51}}%
\newcommand{\xmark}{\ding{55}}%

\newcolumntype{L}{>{\raggedright\let\newline\\\arraybackslash\hspace{0pt}}m{0.45\linewidth}}

\journal{International Journal of Biomedical Informatics}

\begin{document}

\begin{frontmatter}

\title{Automatic Detection of COVID-19 Vaccine Misinformation with Graph Link Prediction}

\author{Maxwell~A.~Weinzierl\corref{cor1}}
\ead{maxwell.weinzierl@utdallas.edu}
\ead[url]{www.utdallas.edu/~Maxwell.Weinzierl}
\author{Sanda~M.~Harabagiu\corref{}}
\ead{sanda@utdallas.edu}
\ead[url]{www.utdallas.edu/~sanda}
\cortext[cor1]{Corresponding author}
\address{Human Language Technology Research Institute, Department of Computer Science, The University of Texas at Dallas, Richardson, TX, USA}

\begin{abstract}

Enormous hope in the efficacy of vaccines became recently a successful reality in the fight against the COVID-19 pandemic. However, vaccine hesitancy, fueled by exposure to social media misinformation about COVID-19 vaccines became a major hurdle. Therefore, it is essential to automatically detect where misinformation about COVID-19 vaccines on social media is spread and what {\em kind} of misinformation is discussed, such that inoculation interventions can be delivered at the right time and in the right place, in addition to interventions designed to address vaccine hesitancy. This paper 
is addressing the first step in tackling hesitancy against COVID-19 vaccines, namely the automatic detection of {\em known} misinformation about the vaccines on Twitter, the social media platform that has the highest volume of conversations about COVID-19 and its vaccines. We present {\sc CoVaxLies}, a new dataset of tweets judged relevant to several misinformation targets about COVID-19 vaccines on which a novel method of detecting misinformation was developed. Our method organizes {\sc CoVaxLies} in a Misinformation Knowledge Graph as it casts misinformation detection as a graph link prediction problem. The misinformation detection method detailed in this paper takes advantage of the link scoring functions provided by several knowledge embedding methods. The experimental results demonstrate the superiority of this method when compared with classification-based methods, widely used currently.

\end{abstract}


\end{frontmatter}

%
%
\section{Introduction}

Enormous hope in the vaccines that inoculate against the SARS-CoV-2 virus, the causative agent of COVID-19, has been building, starting with 2020. When several vaccines have become available, millions signed up and received the vaccines enthusiastically.  
However, too many remain hesitant. Much hesitancy is driven by
misinformation about the COVID-19 vaccines that is spread on social media.
In fact, recent research by \cite{nature-misinfo} has shown
that exposure to online misinformation around COVID-19
vaccines affects intent to vaccinate in order to protect oneself
or others. Therefore, it is essential to automatically detect where misinformation about COVID-19 vaccines on social media is spread and what {\em kind} of misinformation is discussed, such that inoculation interventions can be delivered at the right time and in the right place, in addition to interventions designed to address vaccine hesitancy.

\begin{figure}[]
\centering
\includegraphics[width=0.48\textwidth]{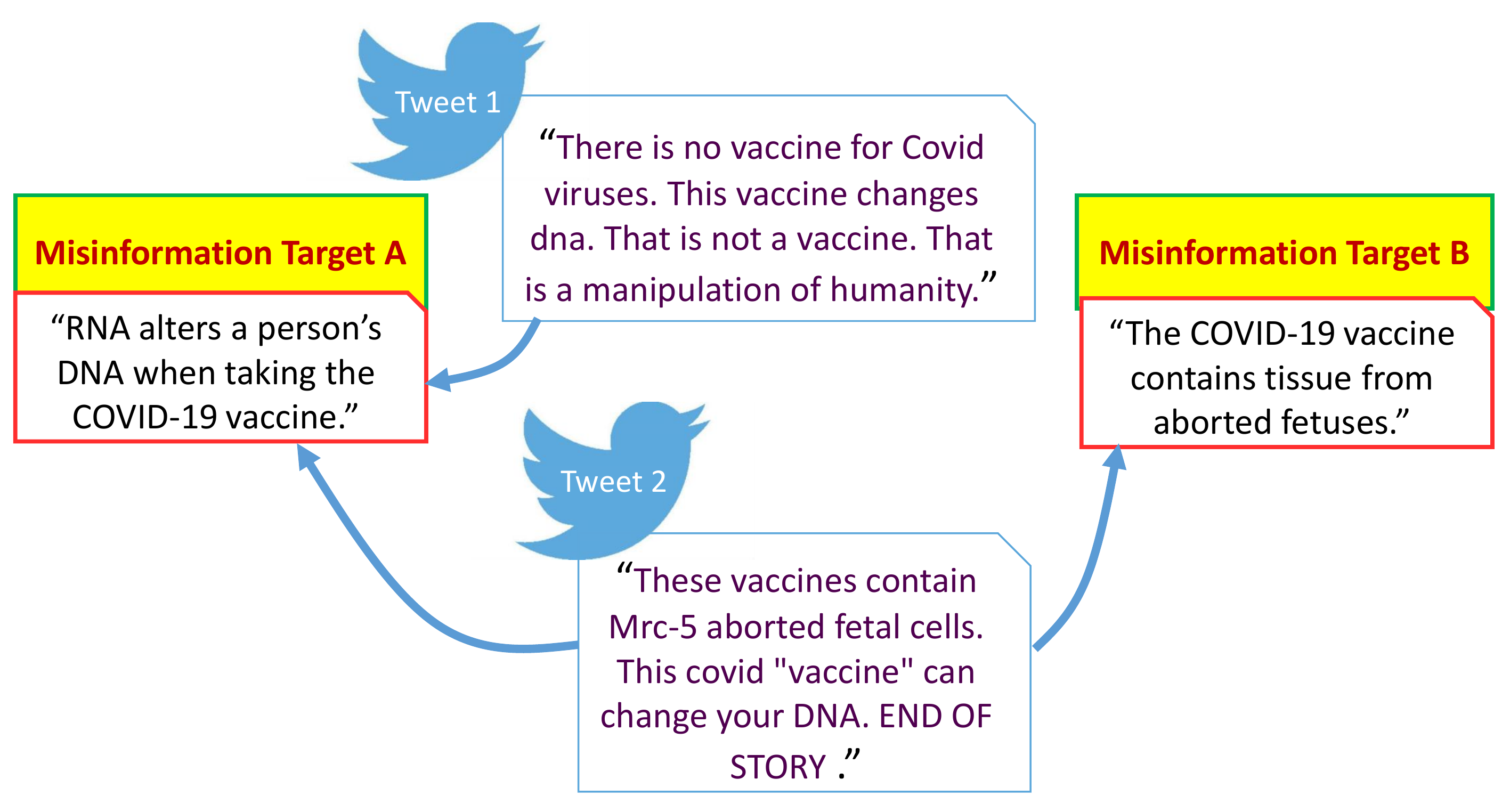}
\caption{Example of tweets containing misinformation about COVID-19 vaccines and the Misinformation Targets referred to by them.}
\label{fig:examples}
\end{figure}

In this paper we address the first step in tackling hesitancy against COVID-19 vaccines, namely the automatic detection of {\em known} misinformation about the vaccines on Twitter, the social media platform that has the highest volume of conversations about COVID-19 and its vaccines. Conversations about the COVID-19 vaccines on Twitter from January to April 2021 were prevalent, with more than 31 million mentions, followed by news, forums, blogs, Reddit, and Tumblr, according to 
research at www.iqvia.com, discussed on one of their blogs \cite{blog-iqvia}. 
The detection of misinformation about COVID-19 vaccines is fundamental in understanding its impact on vaccine hesitancy, by discovering which misinformation is adopted or rejected and how it may influence the attitudes with respect to vaccination.
As with misinformation about COVID-19 in general, there are several misconceptions that are targeted when spreading misinformation about vaccines. These {\sc Misinformation Targets} (MisTs) address commonly {\em known} misconceptions about the vaccines. As illustrated in Figure~\ref{fig:examples}, where two MisTs are illustrated, tweets containing misinformation may be referring to one or multiple MisTs. For example, Tweet$_1$ refers only to MisT$_A$, whereas Tweet$_2$ refers both to MisT$_A$ and MisT$_B$. In order to discover automatically which tweets contain misinformation and to which MisT they refer, we need to design a supervised misinformation discovery method that can be trained on a sufficiently large collection of tweets annotated with misinformation judgements. However, state-of-the-art methods use deep learning techniques, which require a very large training dataset, which is expensive to build. Nevertheless, such a dataset could be bootstrapped from
a seed dataset of high quality when a method of detecting misinformation could operate on it.

In this paper we introduce a tweet dataset annotated with misinformation about COVID-19 vaccines, called {\sc CoVaxLies}, which was inspired by the recently released {\sc COVIDLies} dataset \cite{covidlies},  as well as a method of discovering misinformation on it which can predict links between tweets and MisTs, similar to the links illustrated in Figure~\ref{fig:examples}.
Our framework of discovering misinformation has several novelties. First, it considers that the misinformation about COVID-19 vaccines can be represented
as a Misinformation Knowledge Graph (MKG), in which nodes are tweets that contain misinformation, while edges correspond to the MisTs shared by tweets. Secondly, we propose a representation of the MKG through knowledge embeddings that can be learned by several possible knowledge embeddings models. Thirdly, we use the link ranking functions available from each such knowledge embedding models for predicting a link between any tweet that may contain misinformation and tweets that share a known MisT. Finally, we project the
linguistic content of tweets in the embedding space of the MKG to account not only for the misinformation structure, but also for the language that expressed it. The neural architecture that accounts for all these novelties, a system for Twitter Misinformation Detection through Graph Link Prediction (TMD-GLP) has produced, in our experiments, very promising results on the 
{\sc CoVaxLies} dataset, especially when compared with a neural method that casts misinformation detection as a classification problem, as most current methods do. 

The remainder of the paper is organized as follows. Section 2 describes the related work while Section 3 details the approach used for retrieving tweets relevant to known MisTs regarding COVID-19 vaccines, as well as the expert judgements produced on the relevant data. Section 4 describes our graph-based bootstrapping for misinformation detection and details the neural architecture for Twitter Misinformation Detection through Graph Link Prediction (TMD-GLP). Section 5 presents the experimental results while Section 6 is providing discussions of the results. Section 7 summarizes the conclusions.


\section{Related Work}
\label{sec:related-work}

There are two schools of thought for detecting misinformation on social media, based on (1) the identification of whether a social media posting contains or not misinformation, sometime qualified as a rumour; or (2) taking into account known misconceptions and discovering those postings that propagate a certain misconception. Most of the work belongs to the first school of thought.

{\bf Misinformation Detection as Rumour Identification on Social Media: }Early work aiming the identification of social media postings that contain misinformation (without being interested in its misinformation target) focused on finding useful features for detecting misinformation, e.g. special characters, specific keywords
and expression types, \cite{Castillo}, \cite{Liu}, \cite{Zhao} or  the characteristics of users involved in spreading the misinformation,
e.g. the number of followers, the users’ ages and
genders \cite{Castillo}; \cite{Yang},
and the news’ propagation patterns \cite{Castillo}, \cite{Kwon}. 
More recent work embraced several deep learning methods. These deep learning methods were informed by the textual content of the tweet containing the misinformation, capturing its semantics \cite{Wu},  or by encoding the content of the tweets responding to the misinformation \cite{Ma-ijcai16}. Moreover, a joint recurrent and convolutional network model (CRNN) was reported in \cite{Liu-aaai18}, to better represent the profile of retweeters.
Other deep learning-based methods for the identification of misinformation 
leveraged the propagation
structure in the social network. 
\cite{ma-etal-2017-detect} created
a kernel-based method that captures high-order
interactions differentiating different forms of misinformation while
\cite{ma-etal-2018-rumor}  designed a tree-structured recursive
neural network to learn the embedding of the rumor
propagation structure. Another interesting deep learning framework, reported in \cite{Ma-GAN}, considered the prevalence of
deliberately promoted misinformation campaigns, which can be identified by relying on Generative Adversarial Networks. Most of these misinformation detection methods were influenced by the datasets on which they were developed.

Several well-known benchmark datasets for misinformation detection on Twitter were used previously. For example, the 
Twitter15 \cite{twitter15} and Twitter16 \cite{Ma-ijcai16} datasets consist of a collection of 1490, and 818 respectively, source tweets, along with their corresponding sequences of retweets and replies, forming propagation threads. There were a total of 331,612 propagation threads involving 276,663 users in Twitter15 and 204,820 propagation threads involving 173,487 users in Twitter16. The source tweets were all annotated in balanced sets (of 370 tweets in Twitter15 and 205 tweets in Twitter16) as true rumors, false rumors, unverified rumors or non-rumors. These combined datasets allowed several researchers to develop promising methods trained on the tweets labeled as true or false rumors, while modeling the not only the content of the tweets, but also the retweet/reply sequence
of users, along with user profiles. For example, in \cite{gcan2020} a 
graph-aware representation of user interactions was proposed for detecting 
the correlations between the source tweet content and the retweet propagation though a dual co-attention mechanism. The same idea was explored on the same dataset in the dEFEND system \cite{shu-kdd}. The PHEME dataset \cite{Zubiaga2016} consists of Twitter conversation threads associated with 9 different newsworthy events such as the Ferguson unrest,  the shooting
at Charlie Hebdo, or Michael Essien contracting Ebola. A conversation thread consists of a tweet making a true and false claim, and a series of replies. There are 6,425 conversation threads in PHEME, while only 1,067 claims from tweets were annotated as true,  638 were annotated as false and 697 as unverified. A fraction of the PHEME dataset was used in the RumourEval task \cite{rumoureval}, having only 325 threads of conversations and 145 claims from tweets labeled as true, 74 as false and 106 as unverified. Misinformation detection methods operating both on PHEME and on the RumourEval data sets used either a sifted multi-task learning model
with a shared structure for misinformation and stance detection \cite{sifted}, Bayesian Deep Learning models \cite{baysianDL}
or Deep Markov Random Fields \cite{DeepMRF}.
However, none of these benchmark datasets contain any misinformation about COVID-19 or the vaccines used to protect against it.

{\bf Detection of Known COVID-19-Related Misinformation:}
Very recently, a new dataset of tweets containing misinformation about COVID-19, called {\sc COVIDLies} was released \cite{covidlies}. 
{\sc COVIDLies} is a dataset, which unlike previous datasets  that considered
a large set of ``popular" claims, which were later judged as true, false or unverifiable, was generated by starting with 86 {\em known} misconceptions about COVID-19, available from a Wikipedia article dedicated to misinformation about COVID-19. The misconceptions 
informed the retrieval of 6761 related tweets from COVID-19-related tweets identified by Chen et al\cite{ferrara}. 
The retrieved tweets were further annotated by researchers from the University of California, Irvine School of Medicine with stance information, reflecting their judgement whether the
author of the tweet agreed with a given misconception, rejected the misconception, or the tweet had no stance. Furthermore, {\sc COVIDLies}  enabled the design of a system that could identify misinformation and also infer its stance through a form of neural entailment, as reported in \cite{covidlies}. We also used the {\sc COVIDLies} dataset in recent work to infer automatically when misinformation about COVID-19 is rejected or adopted, by automatically discovering the stance of each tweet against the 86 available misconceptions, which were organized in a taxonomy of misconception themes and concerns. When using
a neural architecture that benefits from stacked Graph Attention
Networks (GATs) for lexico-syntactic, semantic and emotion information, we have obtained state-of-the-art results for stance detection on this dataset, as we report in \cite{our-stance}.

We were intrigued and inspired by the {\sc COVIDLies} dataset, and believed that if we could create a similar dataset containing misinformation about COVID-19 vaccines, which would not only complement the {\sc COVIDLies} data,
but it would also enable the development of novel techniques for misinformation detection. Therefore, in this paper we present the {\sc CoVaxLies} dataset as well as a novel methodology of automatically detecting misinformation using it. We deliberately decided to generate the {\sc CoVaxLies} dataset using a similar methodology as the one employed in the creation of the {\sc COVIDLies} dataset, namely by starting with misconceptions or myths about the vaccines used to immunize against COVID-19 available on a Wikipedia article dedicated to them. But, we cast the misinformation detection problem differently. We still considered the retrieval phase essential for finding relevant tweets for the known vaccine myths, but we explored two different retrieval methods: one using the classic, BM25 \cite{bm25} scoring function, and the other using the same neural scoring method that was used in the creation of {\sc COVIDLies}. This allowed us to discover that classical scoring functions outperform scoring functions using BERT-informed methods. We then focused on producing high-quality judgements for 7,246 tweets against 17 Misinformation Targets (MisTs) about COVID-19 vaccines of interest. Once the {\sc CoVaxLies} dataset was generated, we were able to design a novel, simple and elegant method for discovering misinformation which was cast as learning to predict links in a
Misinformation Knowledge Graph.  Although our method for automatically detection misinformation in a collection of tweets uses deep learning techniques, as most of the recent approaches, it is the first method that 
represents misinformation as a knowledge graph, which can be projected in an embedding space through one of several possible knowledge embedding models.

\begin{table*}[ht]
\centering
\small
\begin{tabular}{p{0.02\linewidth} | p{0.50\linewidth} | p{0.3\linewidth}}
    \toprule
    {\bf ID} & {\bf Misinformation Target } & {\bf Information Source } \\
    \toprule 
    1 & RNA alters a person's DNA when taking the COVID-19 vaccine. & Wikipedia (BBC) \cite{source-wikipedia} \\
    \hline
    2 & The COVID-19 vaccine causes infertility or miscarriages in women. & Wikipedia (Science-Based Medicine) \cite{source-wikipedia}\\
    \hline
    3 & Natural COVID-19 immunity is better than immunity derived from a COVID-19 vaccine. & UC Davis Health \cite{source-ucdavis}\\
    \hline
    4 & The COVID-19 vaccine causes Bell's palsy. & Wikipedia (Snopes) \cite{source-wikipedia}\\
    \hline
    5 & The immune system overreacts to COVID-19 after taking the COVID-19 vaccine through antibody-dependent enhancement. & Wikipedia (Health Feedback) \cite{source-wikipedia}\\
    \hline
    6 & The COVID-19 vaccine contains tissue from aborted fetuses. & Wikipedia (Snopes) \cite{source-wikipedia}\\
    \hline
    7 & The COVID-19 vaccine was developed to control the general population either through microchip tracking or nanotransducers in our brains. & Mayo Clinic \cite{source-mayo1}\\
    \hline
    8 & More people will die as a result of a negative side effect to the COVID-19 vaccine than would actually die from the coronavirus. & Mayo Clinic \cite{source-mayo1}\\
    \hline
    9 & There are severe side effects of the COVID-19 vaccines, worse than having the virus. & Mayo Clinic \cite{source-mayo1}\\
    \hline
    10 & The COVID-19 vaccine is not safe because it was rapidly developed and tested. & Mayo Clinic \cite{source-mayo1}\\
    \hline
    11 & The COVID-19 vaccine can cause COVID-19 because it contains the live virus. & University of Missouri Health Care \cite{source-missuhc} \\
    \hline
    12 & The COVID-19 vaccine causes people to test positive for COVID-19. & University of Missouri Health Care \cite{source-missuhc}\\
    \hline
    13 & The COVID-19 vaccine can increase risk for other illnesses. & University of Missouri Health Care \cite{source-missuhc}\\
    \hline
    14 & Many people already have died from the COVID-19 vaccine trials. & University of Alabama at Birmingham \cite{source-alabu}\\
    \hline
    15 & The COVID-19 vaccine can cause autism. & University of Alabama at Birmingham \cite{source-alabu}\\
    \hline
    16 & The COVID-19 vaccine should not be taken by people who are allergic to eggs. & Mayo Clinic \cite{source-mayo2} \\
    \hline
    17 & Vaccines contain unsafe toxins such as formaldehyde, mercury or aluminum. & PublicHealth.org \cite{source-pubhealth} \\
    
    \bottomrule
\end{tabular}
\caption{COVID-19 {\sc Misinformation Targets} and their sources of information.}
\label{tb:misinfo}
\end{table*}

\section{Retrieving Misinformation about COVID-19 Vaccines from Twitter}
\subsection{COVID-19 Misinformation Targets}

Misinformation about the COVID-19 vaccines has propagated widely, and has been shown to decrease vaccination intent in the UK and USA \cite{nature-misinfo}. 
Medical, public health, and news organizations have scrambled to collect and rebut misinformation surrounding the COVID-19 vaccine in response to this spread. 
%
Organizations such as the Mayo Clinic, University of Missouri Health Care, University of California (UC) Davis Health, University of Alabama at Birmingham, Science-Based Medicine, PublicHealth.org, Snopes, and the British Broadcast Corporation (BBC) have been actively collecting misinformation about the COVID-19 vaccines and debunking them on public websites. 
Wikipedia also has an entire page dedicated to COVID-19 misinformation, which collects many misconception claims, including those referring to the vaccines developed for immunization against the COVID-19 virus. The Wikipedia page available at {\em en.wikipedia.org/wiki/COVID-19\_misinformation\#Vaccines}  also provides citations to scientific articles that debunk the misconceptions. 
For example, the BBC is cited on the Wikipedia page above mentioned for identifying and debunking the misconception that ``RNA alters a person’s DNA when taking the COVID-19 vaccine." 
In the cited article \cite{bbc-misinfo} the authors claim that ``The fear that a vaccine will somehow change your DNA is one we've seen aired regularly on social media." 
They immediately debunk this misinformation by claiming: ``The BBC asked three independent scientists about this. They said that the coronavirus vaccine would not alter human DNA."

We took advantage of the existing efforts of pinpointing the misconceptions related to the COVID-19 vaccines and debunking them. We selected
17 misinformation claims, which we considered as {\sc Misinformation Targets}, because the propagation of misinformation on social media targets one or several such misconceptions. In Table~\ref{tb:misinfo} we list all the {\sc Misinformation Targets} (MisTs) that we considered in this study along with their source of information. 

\subsection{Retrieval Methods for Identifying Tweets with Potential Misinformation Content}
\label{sec:data-ir}
Before using the Twitter streaming API to collect tweets discussing the COVID-19 vaccine, approval from the Institutional Review Board at the University of Texas at Dallas was obtained: IRB-21-515 stipulated that our research met the criteria for exemption \#8(iii) of the Chapter 45 of Federal Regulations Part 46.101.(b). 
When using the search query ``(covid OR coronavirus) AND vaccine" on the Twitter streaming API, we collected 840,299 English tweets and we ignored all retweets. 
{Our investigation of the Twitter platform revealed that} Twitter’s {tokenization of tweets splits up terms like}
``covid19” and ``covid-19” into ``covid” and ``19”. 
{Therefore we selected} ``covid” 
{as a search term that matches the tokenization not only of mentions of "covid" in tweets, but also mentions of "covid-19" or "covid19", thus optimizing the recall of relevant tweets, when combined with the keywords} ``coronavirus" and ``vaccine".
The retrieved tweets were authored in the time frame from December 18th, 2019, to January 4th, 2021.
A large fraction of these tweets were duplicates, likely due to spam bots, which required filtering. 
Locality Sensitive Hashing (LSH) \cite{lsh} is a well-known method used to remove near-duplicate documents in large collections. 
We perform LSH, with term trigrams, 100 permutations, and a Jaccard threshold of 50\%, on our collection to produce $\cal{C}_{T}$= 753,017 unique tweets.

We found that approximately 35\% of the unique tweets in $\cal{C}_{T}$ referred to an external news article, YouTube video, blog, or other website. 
Therefore, we also crawled these external links and parsed their contents with Newspaper3k \cite{newspaper3k} to include their titles with the original tweets. We found these titles added significantly more context to many of these tweets, and allowed us to identify many more instances of misinformation during the human judgement phase. For example, the tweet {\em ``Who still want to be vaccinated? URL: ``Vaccine causing bells palsy: Pfizer vaccine side effects, covid"} would be impossible to identify as pertaining to MisT$_4$ without the article's context.

In order to identify in $\cal{C}_{T}$ those tweets which potentially contain language relevant to the MisTs of interest, listed in Table~\ref{tb:misinfo}, we relied on two information retrieval systems: (1) a retrieval system using the BM25 \cite{bm25} scoring function; and (2) a retrieval system using  BERTScore \cite{bertscore} with Domain Adaptation (DA), identical to the one used in \cite{covidlies}. 
Both these retrieval systems operated on an index of $\cal{C}_{T}$, obtained by using Lucene \cite{lucene}. Each retrieval system produced a ranked list of tweets when queried
with the textual content of any of the MisTs of interest, listed in Table~\ref{tb:misinfo}.
At most 200 top scored tweets were selected for each of these queries. 
We selected only 200 best scored tweets because (1) the same number of tweets was considered in the most similar prior work \cite{covidlies}; and (2) it was a number of tweets that did not overwhelm our human judges. 
Additional tweets were also retrieved when replacing the word ``COVID-19" with the word ``coronavirus" in each query.  From the top ranked tweets deemed relevant to the modified query, at most 200 tweets were also considered. This approach produced a maximum of 400 tweets, but the list of retrieved tweets returned by the retrieval systems often contained less than 400 tweets. For example, when MisT$_{10}$ was used as a query, we were able to retrieve only 213 distinct tweets, whereas when MisT$_{11}$ was used as a query, we retrieved 282 distinct relevant tweets. Some of the tweets retrieved when the query that was used was one of the MisTs listed in Table~\ref{tb:misinfo} were also retrieved when the modified query was used.
In the end, we retrieved a total of 4,153 tweets that were deemed relevant to at least one MisT of interest.

When using the retrieval system which relies on the BERTScore (DA), we were able to benefit from its domain-adaptive pre-training on 97 million COVID-19 tweets. In this way, semantic relevancy to the domain of COVID-19 was preferred over keyword matching when scoring the tweets against the MisTs of interest. As with the first retrieval system, at most 200 top scored tweets were selected for the original queries and at most other 200 tweets when the modified queries, 
replacing the word ``COVID-19" with the word ``coronavirus", were used. In this way, the second retrieval system enabled us to collect 4,689 tweets deemed relevant to at least one MisT of interest. Obviously, the second retrieval system enabled us to collect a larger set of tweets deemed relevant to the 
MisTs of interest than the first retrieval system (4,689 tweets vs. 4,153 tweets), which we attribute to significant sensitivity in BERTScore when replacing the word ``COVID-19" with ``coronavirus" in the query. 
As in the case when the retrieval system using the BM25 scoring function, 
some of the tweets retrieved by the retrieval system using when BERTScore (DA), when the query was one of the MisTs listed in Table~\ref{tb:misinfo}, were also retrieved when the modified query was used. This time, the fewest distinct tweets were returned for the MisT$_4$ (only 234 distinct tweets), whereas when the query used the MisT$_3$, we collected 332 distinct tweets. 
We finally combined the collection of unique tweets retrieved with both systems, 
obtaining a set $\cal{T}_R$ of 7,246 tweets deemed relevant to our MisTs of interest. {A total of 1,596 tweets from $\cal{T}_R$ were considered relevant to the same query (any MisT) by both retrieval systems.} It is to be noted that some tweets from $\cal{T}_R$ were relevant to more than one of the MisTs listed in Table~\ref{tb:misinfo}.

\begin{figure*}[t]
\centering
\includegraphics[width=0.80\textwidth]{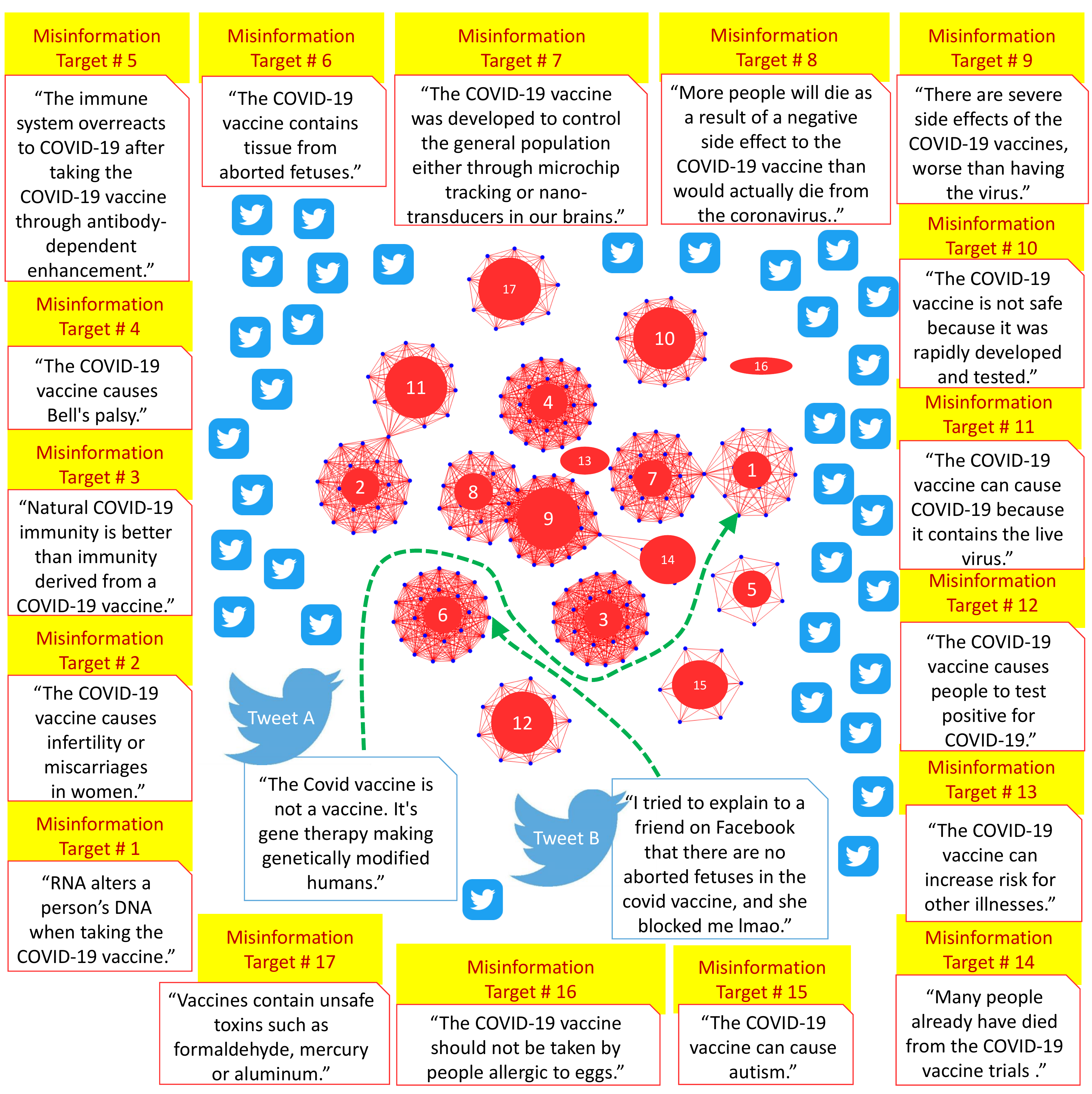}
\caption{Misinformation Detection as Graph-Link Prediction}
\label{fig:graph}
\end{figure*}

\subsection{Producing Human Judgements for the Misinformation Contained in Tweets}
\label{sec:data}

Given each tweet from the $\cal{T}_R$ collection of 7,246 tweets deemed relevant to our MisTs of interest, researchers from the Human Language Technology Research Institute (HLTRI) at the University of Texas at Dallas produced judgements of either: {\em Relevant} or {\em Not\_Relevant} against the MisTs listed in Table~\ref{tb:misinfo}. Each tweet was read and when it contained a discussion about one of the MisTs of interest, it was considered relevant.
It is to be noted that sometimes, the tweet could discuss more than one MisT, therefore it was judged relevant to each MisT it addressed. 
{A total of 836 tweets from $\cal{T}_R$ were judged relevant to two or more MisTs.}
Alternatively, when the tweet did not discuss any MisT of interest, it was judged non-relevant to any of the MisTs. It should be noted that, when a tweet was judged relevant, no additional judgements about the stance of the tweet with respect to the MisT were considered, disregarding if the tweet agreed, disagreed with the MisT, or had no stance about it. This decision was made because stance detection was regarded as a separate task, as in \cite{covidlies},\cite{our-stance}, which requires each tweet to be first found relevant to the MisT, before inferring its stance. As it is illustrated in Figure~\ref{fig:graph}, Tweet A is relevant to MisT$_1$, agreeing with it, while Tweet B is relevant to MisT$_6$, disagreeing with it.

To evaluate the quality of judgements, 
we randomly selected a subset of 1,000 tweets (along with the MisT against which they have been judged to be relevant or non-relevant), which have been judged by at least three different language experts. Percent agreement between annotators was 92\%. Fleiss' Kappa score was 0.83, which indicates strong agreement between annotators (0.8-0.9) \cite{kappa}. 
There were high levels of agreement between the annotators of tweets against specific MisTs: MisT$_1$, Mist$_2$, MisT$_4$, MisT$_6$, MisT$_7$, MisT$_{10}$, MisT$_{14}$ and MisT$_{17}$. This can be explained by the unique interpretation of these MisTs and their direct reference in tweets. However, the inter-annotator agreement for the other MisTs were lower, with the lowest Fleiss' Kappa score of 0.70 obtained for MisT$_8$, and the highest Fleiss' Kappa score of 0.89 obtained for MisT$_{14}$.
Disagreements in annotations were discussed, but largely came down to interpretation. 
For example, the tweet ``we dont need a covid vaccine" was interpreted by one expert as relevant to the MisT$_3$: ``Natural COVID-19 immunity is better than immunity derived from a COVID-19 vaccine", while  another judge found  the tweet non-relevant to the same MisT. 
For the first annotator, the statement that no vaccine is necessary for COVID-19 implied that the author of the tweet entailed that that natural immunity from catching COVID-19 is better than any vaccine-induced immunity. 
The second judge took a more strict interpretation, recognizing that there may be other reasons that the tweet author might have when stating that a COVID-19 vaccine is unnecessary, unrelated to the MisT$_3$. 
The differences of interpretation were analyzed and resolved, and judges  were instructed to be strict in determining relevant tweets for each MisT of interest. In the end, this collection of judged tweets, which we call
{\sc CoVaxLies}, contains 7,246 [tweet, MisT] {\em pairs}, where each tweet was judged as either relevant or non-relevant against the MisT it is paired with. 

We split the {\sc CoVaxLies} into three distinct collections: (a) a training collection; (b) a development collection; and (c) a test collection. The training collection, which consists of 5,267 [tweet, MisT] pairs, was utilized to train our automatic misinformation detection systems, described in Section~\ref{sec:methods}. {The development collection, which consists of 527 [tweet, MisT] pairs, was used to select model hyper-parameters, such as threshold values, as well as for building the seed of the} {\em Misinformation Knowledge Graph}, described in Section~\ref{sec:graph_link}. {The test collection, which consists of 1,452 [tweet, Mist] pairs, was used to evaluate misinformation detection approaches, described in} Section~\ref{sec:results}.

\section{Graph-Based Bootstrapping of Misinformation Detection}
\label{sec:methods}

\subsection{Misinformation Detection as Graph-Link Prediction}
\label{sec:graph_link}
The way misinformation spreads on social media platforms inspired us to consider that misinformation can be represented as a {\em Misinformation Knowledge Graph} (MKG), in which nodes represent tweets, containing misinformation about COVID-19 vaccines, while links between tweets indicate that they share the same misinformation, and thus are labeled with the corresponding {\sc Misinformation Target} (MisT). 
Therefore, for each MisT a separate, fully-connected graph (FCG) can be generated, spanning all tweets that discuss that MisT. Some tweets may be connected to multiple MisT-specific FCGs, if they contain information pertaining to more than one MisT. Furthermore, tweets with text that is not relevant to any of the MisTs of interest shall remain as unconnected nodes in the MKG.  
Figure~\ref{fig:graph} illustrates visualizations of the MKG with FCGs pertaining to several MisTs, each labeled with a different number, while the content of the MisT labeled with the same number is illustrated in the Figure as well.
For example, the FCG(MisT$_4$)  includes as nodes tweets which are relevant to the MisT$_4$:  ``The COVID-19 vaccine causes Bell's palsy." The edges in FCG(MisT$_4$) are all labeled as MisT$_4$. It is to be noted that in Figure~\ref{fig:graph} some tweets are connected to multiple MisT-informed FCGs.  For example, the tweet {\em ``Omg this! URL QT: ``It makes no sense. You are ok with people getting Covid-19 (a dangerous virus) because they are unlikely to die from it in order to create herd immunity, but not people getting a vaccine that might have some minor side effects for a small portion to create that immunity?"} is relevant to both MisT$_8$ and MisT$_9$, and thus is represented as a node participating in both FCG(MisT$_8$) and FCG(MisT$_9$).

The goal of identifying MisT-specific misinformation is achieved by generating and bootstrapping the MKG for a collection of tweets, such as {\sc CoVaxLies}, in two steps:\\
\underline{STEP 1:} {\em Generate the seed FCG for each MisT of interest.}
The seed FCGs, which are illustrated in Figure~\ref{fig:graph}, were informed by the development set of {\sc CoVaxLies},  described in Section~\ref{sec:data}. Whenever two tweets were judged {\em Relevant} to the same MisT, they were connected with a link labeled by that MisT. In this way, all tweets judged relevant to the same MisT become nodes in an FCG corresponding to that MisT. Some of the MisT-informed FCGs are larger than others, because there are more tweets relevant to that MisT in the development data from {\sc CoVaxLies}. 
The largest FCG has 27 tweets, all pertaining to MisT$_9$ while the smallest FCG has only 1 tweet pertaining to MisT$_{16}$. 
From the total of 527 [tweet, MisT] pairs available in the development set of {\sc CoVaxLies}, in this step, a total of 245 tweets were assigned to all 17 FCGs, while  170 tweets were left unconnected because they were  judged not to be relevant to any MisT. The number of tweets either connected or not connected to any FCG is smaller than the number of pairs from the development set of {\sc CoVaxLies} because some tweets were relevant to multiple MisTs. \\
\underline{STEP 2:} {\em Bootstrap the seed MisT-specific FCGs to span additional tweets sharing the same misinformation target}.
The bootstrapping is enabled by {\em learning} to predict if any link(s) can be established between any of the FCGs and tweets from a collection of tweets. In our experiments, we shall consider {\sc CoVaxLies} as the collection of tweets, since {\sc CoVaxLies} is already split into a training collection, a development collection and a testing collection, enabling a machine learning framework. The development collection was used to create the seeds of the FCGs, in Step 1. In Step 2, we learn to predict links to the FCGs by considering the training collection of {\sc CoVaxLies}, and then we test the quality of predictions on the testing collection of {\sc CoVaxLies}. This entails that the bootstrapping has two phases, the first phase relying on the training collection, while the second phase relying on the testing collection of {\sc CoVaxLies}.

In Phase 1 of the bootstrapping of the MKG, the training collection of {\sc CoVaxLies}, comprising 5,267 [Tweet, MisT] pairs is used to bootstrap the seed FCGs that were generated in Step 1. For each [Tweet$_x$, MisT$_y$] pair, a link is established between the tweet$_x$ and the FCG(MisT$_y$), providing a training example for learning to predict links between other, unconnected tweets and the various MisT-specific FCGs from the MKG. A consequence of linking any Tweet$_x$ to the FCG(MisT$_y$) is that a node for Tweet$_x$ is added to FCG(MisT$_y$), along with all the MisT$_y$-labeled edges between 
Tweet$_x$ and the remaining nodes from FCG(MisT$_y$), such that FCG(MisT$_y$) remains a fully connected graph after incorporating Tweet$_x$. There are only {3,735} unique tweets in the training collection of {\sc CoVaxLies}, because many of them were linked to more than one MisT-specific FCG. 
Therefore, in Phase 1 of Step 2, in our experiments, we added initially 5,267 links to the FCGs generated in Step 1, which enabled the addition of {2,114} tweets to the FCGs produced in Step 1. 

In Phase 2 of the bootstrapping of the MKG, the test set of {\sc CoVaxLies}, comprising  1,452 [Tweet, MisT] pairs, is considered. However, there are only 1,038 unique tweets in the test collection of {\sc CoVaxLies}, as many of these tweets were judged to be relevant to multiple MisTs. All these 1,038 tweets are initially {\em unconnected} to any MisT-specific FCGs, because knowledge about their relevance 
to any of the MisTs is withheld, such that the evaluation of the link predictions can be performed. 
In Phase 2, bootstrapping of MisT-specific FCGs is performed by predicting if a link exists from an unconnected tweet to any FCG. When a link is predicted, misinformation is discovered, and moreover, its 
misinformation target is also discerned. If no link is predicted, then no misinformation pertaining to any of the seventeen MisTs of interest is discovered. 
In this way, we cast the problem of misinformation detection as a {\bf graph link prediction} problem. For example, given the $Tweet_A$, illustrated in Figure~\ref{fig:graph}, which was initially unconnected to any FCG in the MKG,  we learn to detect that it contains misinformation characteristic of MisT$_1$, based on the training data produced in the Phase 1 of Step 2 of bootstrapping. Consequently, a link is predicted between $Tweet_A$, and the FCG(MisT$_1$). Similarly,  we learn to predict a link between $Tweet_B$ and the FCG(MisT$_6$), thus discovering new misinformation sharing that MisT. As link predictions are performed at test time, in Phase 2 of the bootstrapping of the MKG, tweets are connected to one or multiple of the MisT-informed FCGs and the final MKG emerges. Moreover, Phase 2 of the Step 2 of bootstrapping enables us to evaluate the quality of the link prediction models that we have considered, while Phase 1 of the Step 2 enabled us to train these models on the training collection of  {\sc CoVaxLies}. It is to be noted that the MKG can be bootstrapped by considering all the 753,017 tweets from $\cal{C_T}$ which do not belong to {\sc CoVaxLies}, however, the quality of the link prediction in 
Phase 2 of the Step 2 could not be evaluated, as no misinformation relevance judgements are available for those tweets.

When formalizing the MKG as a knowledge graph  $MKG=(V;E)$, the tweets are 
represented as
$t_i \in V$, where $V$ represents the entire collection of tweets from {\sc CoVaxLies}, while $E$ represents the misinformation links that might exist between these tweets. For example, all tweets that were linked to an FCG(MisT$_y$) (either in Step 1 or Step 2 of the bootstrapping of the MKG) will share 
edges labeled MisT$_y$, shared by each pair of tweets. This formalization of the MKG enables us to learn
to predict links between unconnected tweets and the FCGs from the MKG.

For more than a decade, knowledge graphs were represented in a continuous vector space called the {\em embedding space} by learning {\em knowledge embeddings} for its nodes and edges.  Given the MKG, knowledge embedding models learn an embedding $te_i$ for each tweet $t_i$ and an embedding $me_j$ for each MisT$_j$, with $j$ ranging over all MisTs used as labels in the MKG. More importantly,  knowledge embedding models use a {\em link scoring function} 
for assigning a plausibility score to any potential link between two tweets of the MKG, $t_i$ and $t_k$, relying on the embeddings of the pair of tweets as well as on the embedding learned for any MisT$_j$ : $f(te_i,me_j,te_k)$.  Because knowledge embedding models provide a link scoring function, we adopted them as an ideal framework for resolving the problem of graph link prediction and apply it to misinformation detection in tweet collections.
In Section~\ref{sec:ke-models} we discuss several knowledge embedding models and provide their link scoring functions.

However, the link scoring function is necessary but not sufficient for bootstrapping the MKG in Phase 2 of Step 2. When deciding to which FCG an unconnected tweet could be linked, we are presented with two options, each yielding different computational complexities for computing the link scoring function:\\
{\em Option 1}: Select all of the tweets from any FCG;  or\\
{\em Option 2} Select a prototypical tweet from each FCG. Because the nodes of each FCG are represented by their knowledge embeddings, thus vectors, the centroid of these embeddings can be cast as the prototypical representation of each FCG.  Thus the representation of the prototypical tweet for 
each FCG needs to take into account the knowledge embeddings of all the tweets connected to the FCG. Therefore, for any FCG(MisT$_x$) having $n_x$ nodes, corresponding to its tweets $t^{x}_i$, encoded by the knowledge embedding ${te^{x}_i}$, the representation of the prototypical tweet of FCG(MisT$_x$) is  provided by its embedding $pte^{x}$, computed as: 
\begin{equation}
    pte^{x}=\frac{1}{n_x}\sum_{k=1}^{n_x}{{te^{x}_k}}
\end{equation}
When considering option 1, the link scoring function is computed between a new, unconnected tweet $t^{y}$
and each of the $n_x$ tweets $t^{x}_i$ of each FCG(MisT$_x$). Then, the {\em Condition}$_{ALL}$ must be satisfied, where {\em Condition}$_{ALL}$ stipulates that if more than $N_x$ number of times the value returned by the link prediction function is superior to a threshold $T_x$, the link is predicted between $t^{y}$ and FCG(MisT$_x$). Clearly, the number of tweets $n_x$, and $N_x$ and $T_x$ are dependent on each MisT$_x$, varying across MisTs. 
{We assign } $N_x$ and $T_x$ {automatically by maximizing misinformation detection performance of the system for MisT$_x$ on the development collection, which is further detailed in} Section~\ref{sec:eval-misinfo}.
However, the number of times the link scoring function
$f$ needs to be computed when considering this option is equal to the number of tweets that are already connected in the MKG at the time of attempting to link a new tweet. This number easily grows in the thousands - and thus it renders this options computationally inefficient. 

The option 2 presents the advantage that the link prediction function $f$ is evaluated only once for each MisT encoded in the MKG, and a link is predicted when the value returned by $f$ is superior to a pre-defined threshold for each MisT, $T_x$. 
Moreover, in both options, different link scoring functions, available from different knowledge embedding models, may predict different graph links in the MKG, and thus discover differently the misinformation in a collection of tweets.

Our misinformation detection framework using graph link prediction allows for rapid experimentation with multiple knowledge embedding models to explore their performance. Each of them enables the learning of misinformation knowledge embeddings, in the form of knowledge embeddings of the tweets that may contain misinformation, as well as knowledge embeddings for each MisT of interest.


\begin{figure*}[t]
    \centering
    \includegraphics[width=0.60\textwidth]{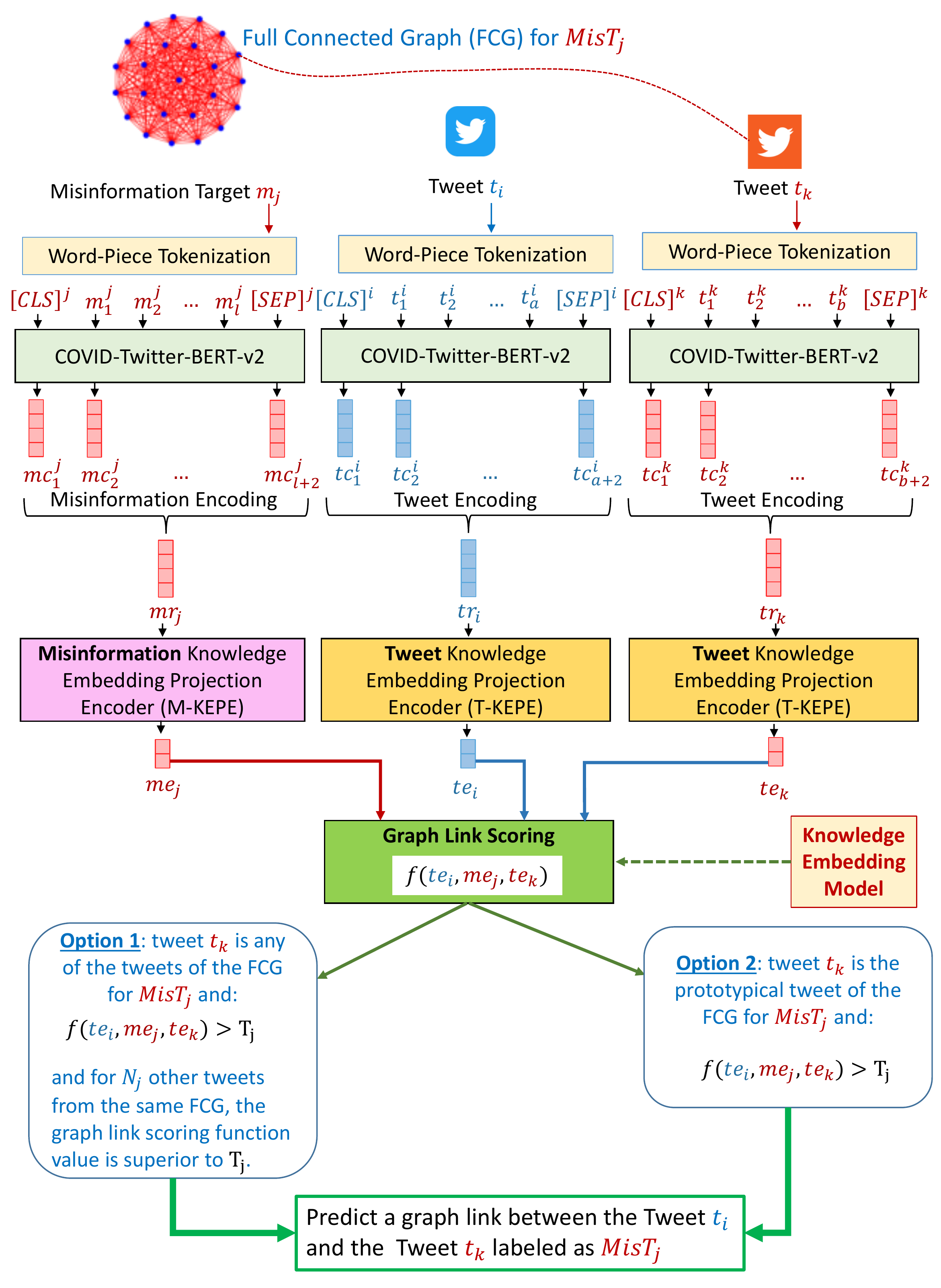}
    \caption{Neural Architecture for Twitter Misinformation Detection through Graph Link Prediction (TMD-GLP).}
    \label{fig:architecture}
\end{figure*}

\subsection{Learning Misinformation Knowledge Embeddings}
\label{sec:ke-models}

Several knowledge embedding models have been widely used in the past decade, e.g. TransE~\cite{transe}, TransD~\cite{transd}. 
In addition to TransE and TransD,  several other knowledge graph embedding models have shown promise in recent years, e.g. TransMS~\cite{transms} and TuckER~\cite{tucker}. We have explored how all these four different knowledge embedding models perform in our framework for misinformation detection as graph link prediction. We briefly describe them before discussing how they were used in a novel neural architecture for twitter misinformation detection as graph link detection.

{\bf TransE}, like all knowledge embedding models, learns an embedding, for each node in the knowledge graph and an embedding for each relation type. Given the MKG, TransE learns an embedding ${te_i}$ for each tweet $t_i$ and an embedding ${me_j}$ for each MisT$_j$, with $j$ ranging over all MisTs represented in the MKG. Moreover, TransE considers that the link embedding is a {\em translation vector} between the two tweet embeddings representing its arguments.
This means that for any tweet $t_i$, the tweet most likely to be linked to $t_i$ by a link labeled MisT$_j$ should be the tweet whose embedding is closest to $(te_i + me_j)$ in the embedding space.
By modeling the tweets as points in the embedded space and the MisTs they share as translation vectors, it is possible to measure the \textit{plausibility} of any potential link labeled as MisT$_j$ between any pair of tweets $t_i$ and $t_k$ using the geometric structure of the embedding space: 
\begin{equation}
f(te_i, me_j, te_k) = -||te_i + me_j - te_k||_{L1}
\end{equation}
where $||\cdot||_{L1}$ is the $L1$ norm.
The plausibility of a relation between an unconnected tweet $t_u$ and tweet $t_d$ connected to (or prototypical of a) FCG-x with labels MisT$_x$, represented as a triple, $\langle t_u, MisT_x, t_d \rangle$, is inversely proportional to the \textit{distance} in the embedding space between the point predicted by the TransE model, ($te_u + me_x$), and the point in the embedding space representing the destination argument of the link labeled as MisT$_x$, i.e. ($te_d$).
TransE has the advantage that it is extremely simple to utilize, but interactions between node embeddings and edge embeddings are limited. 

{\bf TransD} extends TransE by learning two knowledge embeddings for each node and each edge from a knowledge graph
such that the first embedding represents the ``knowledge meaning" of the node or relation while the second embedding is a \textit{projection} vector (with superscript p), used to construct a dynamic mapping matrix for each node/link pair. Thus, for each tweet from the MKG,
$t_i$ TransD learns the pair of embedding $(te_i,te^p_i)$ and for each link labeled as MisT$_j$, it learns the pair of embeddings $(me_j,me^p_j)$. The pair of knowledge embeddings for the tweet and for the link are learned by using a scoring function that measures the plausibility of a link labeled MisT$_j$ between a tweet $t_i$ and a tweet $t_k$,  defined as:
\begin{equation}
\begin{split}
    f(te_i,me_j,te_k)= 
    -\left\lVert (\bm{I} + me^p_j \times ({te^p_i})^{\top}) \times te_i + me_j \right. \\ 
    - \left. (\bm{I} + me^p_j \times ({te^p_k})^{\top} ) \times te_k \right\rVert_{L1} 
\end{split}
\end{equation}
where $\bm{I}$ is the identity matrix. 

TransD improves upon TransE by modeling the interactions between tweets and the links that span them through their respective knowledge embeddings, such that tweet embeddings change depending on which MisT is being considered for labeling a link. 

{\bf TransMS} recognizes the importance of capturing non-linear interactions between nodes and edges in a knowledge graph, and therefore expands on the approach of TransD. TransMS introduces non-linear interactions on both the node and edge knowledge embeddings before the additive translation of TransE is performed, and also adds an edge-specific threshold parameter $\alpha_j$.
When considering the MKG, the knowledge embeddings for the tweet and for the link are learned by using a scoring function that measures the plausibility of a link labeled MisT$_j$ between a tweet $t_i$ and a tweet $t_k$, defined as:
\begin{equation} \label{eq:transms}
\begin{split}
    f(te_i,me_j,te_k)=-
    \left\lVert - tanh(te_k \otimes me_j) \otimes te_i 
    + me_j \right. \\
    + \left. \alpha_j \cdot (te_i \otimes te_k) 
    - tanh(te_i \otimes me_j) \otimes te_k \right\rVert_{L1}
\end{split}
\end{equation}
where $tanh(x)$ is the non-linear hyperbolic tangent function and $\alpha_j$ is a real numbered parameter dependent on each MisT. The operator $\otimes$ represents the Hadamard product. 

TransMS improves upon TransD by allowing both the nodes to influence the edge embeddings and the edges to influence the node embeddings. TransMS also introduces non-linearities in these interactions, and allows edge type-specific $\alpha_j$ thresholds to be learned.  

{\bf TuckER} utilizes a multiplicative approach to learning knowledge embeddings, as opposed to the additive approach of TransE, TransD, and TransMS. TuckER introduces a learned ``core tensor" $\mathcal{W} \in \mathbb{R}^{z \times v \times z}$ which encodes some of the learned knowledge outside of the knowledge embeddings. When considering the MKG, the knowledge embeddings for tweets and for links spanning them are learned by using a scoring function that measures the plausibility of a link labeled MisT$_j$ between a tweet $t_i$ and a tweet $t_k$. The pair of knowledge embeddings for the tweet and for the link are learned in TuckER by using a scoring function that measures the plausibility of a link labeled MisT$_j$ between a tweet $t_i$ and a tweet $t_k$,  defined as:
\begin{equation}
\begin{split}
    f(te_i,me_j,te_k)= \mathcal{W} \times_1 te_i \times_2 me_j \times_3 te_k
\end{split}
\end{equation}
where $\times_n$ indicates the tensor product along the n-th mode. TuckER approaches the problem of learning knowledge embeddings from a multiplicative perspective, with an additional component in the $\mathcal{W}$ tensor which allows for additional shared interactions to be learned between the nodes and edges of the knowledge graph through tensor products. 

In addition to the knowledge embedding models, we also considered a K-Nearest Neighbors (KNN) baseline approach. The KNN approach ignores entirely the edge information available in the FCC(Mist$_j$) of each Mist$_j$. Instead, this approach favors tweets that are closest in their representation in the embedding space. Thus, when scoring a link between an unconnected  tweet $t_i$, represented by a knowledge embedding $te_i$, and any tweet $t_k$, from the FCC(MisT$_j$),  represented as $te_k$, it computes:
\begin{equation}
f(te_i, me_j, te_k) = -||te_i - te_k||_{L2}
\end{equation}
where $||\cdot||_{L2}$ is the $L2$ norm. To predict the link to any MisT of interest, e.g. MisT$_x$, the
{\em Condition}$_{ALL}$, defined in Section~\ref{sec:graph_link}, must be met. This condition is applied because all nodes that are already assigned to all FCGs from the MKG need to be considered when searching in the embedding space for the tweets that are closest to $t_i$. For the KNN approach, this condition 
requires that 
there must be at least $N_x$ values of $f$ which are larger than a given threshold $T_x$ for all nodes $n_x$ connected in the FCG(Mist$_x$). When the condition is satisfied, a link is predicted between 
$t_i$ and MisT$_x$. But the condition may be satisfied by more than one MisT of interest, and thus multiple links may be predicted. {This baseline is equivalent to a KNN approach, where the distance metric is the Euclidean distance, and the value of $K$ is selected automatically based on (a) the number of MisTs for which {\em Condition}$_{ALL}$ is satisfied, which implicitly depends on (b) the development collection available from} {\sc CoVaxLies}, which was used to generate the seed FCGs from the 
MKG.

Any of the scoring functions of these four knowledge embedding models or the baseline model using KNN can be used to predict a graph link in the MKG. But, because the tweets use natural language to communicate their message, and each of the MisTs also are expressed in natural language, the subtleties and deep connections expressed in language also need to be captured. The language used in two different tweets from the MKG, that might be linked together need to accounted for. In addition, the language describing the MisT they share needs also to be accounted for when predicting if there is a link. With this in mind, we have designed a neural architecture for Twitter Misinformation Detection through Graph Link Prediction (TMD-GLP), illustrated in Figure~\ref{fig:architecture}.

Given a MisT $m_j$, the TMD-GLP system first performs Word-Piece Tokenization \cite{bert} on the textual description of the misinformation target, producing tokens $m_1^j, m_2^j, ..., m_l^j$, where $l$ is the number of word-piece tokens produced.
Word-Piece Tokenization segments text into word-pieces, where the segmentation is data-driven and based on maximizing the size of frequently occurring sub-word tokens.
Common English words remain intact as whole tokens, while uncommon words are segmented into more common sub-word components. 
This process ensures every word can be segmented into a known vocabulary of sub-word units, and no words are left out-of-vocabulary. 
{The $[CLS]$ and $[SEP]$ tokens are placed at the beginning and end of the token sequence respectively, as was done in prior work to denote the beginning and end of the token sequence} \cite{bert}.

The word-piece tokens $[CLS]^j, m_1^j, m_2^j, ..., m_l^j, [SEP]^j$ are then provided to the BERT \cite{bert} COVID-19 Language Model COVID-Twitter-BERT-v2 \cite{covid-twitter-bert}.
COVID-Twitter-BERT-v2 is a pre-trained domain-specific language model, which means that it started with neural weights equal to those of BERT, but was additionally pre-trained on the masked language modeling task \cite{bert} for 97 million COVID-19 tweets. 
This process of further pre-training has been shown to improve performance on downstream tasks in various scientific \cite{scibert}, biomedical \cite{biobert}, and social media \cite{bertweet} domains. 
COVID-Twitter-BERT-v2 therefore produces contextualized embeddings $mc_1^j, mc_2^j, ..., mc_{l+2}^j$ for the word-piece tokens in the MisT $m_j$ along with the $[CLS]^j$ and $[SEP]^j$ tokens.  
In this way, 
we encode the language describing the MisT using a contextualized embedding $mr_j \in \mathbb{R}^{1024}$, where $1024$ is the contextual embedding size for COVID-Twitter-BERT-V2, which is the first contextualized embedding $mc_1^j$, {representing the initial $[CLS]^j$ token embedding}.

Similarly, the language used in the tweets $t_i$ and $t_k$ is processed through Word-Piece Tokenization and then represented by contextual embeddings $tr_i$ and $tr_k$ after being processed through COVID-Twitter-BERT-v2.
But, it is important to note, that the scoring function $f$ of any of the knowledge embedding models that we considered, illustrated in Figure~\ref{fig:architecture}, cannot operate directly on the contextual embeddings $tr_i$, $mr_j$ or $tr_k$, as they do not have the same dimensions of the knowledge embeddings these models learn. Therefore, in TMD-GLP we needed to consider two forms of projection encoders, capable to project from the contextualized embedding space into the knowledge embedding space. For this purpose, we have relied on  the Misinformation Knowledge Embedding Projection Encoder (M-KEPE), using a fully-connected layer to project from $mr_j$ into the necessary knowledge embedding $me_j \in \mathbb{R}^{v}$ of any of the knowledge embedding models considered. Similarly, we have 
relied on the Tweet Knowledge Embedding Projection Encoder (T-KEPE) using a different fully-connected layer than M-KEPE. As shown in Figure~\ref{fig:architecture}, these encoders produce the arguments of the scoring function, which informs the decision whether the link between the unconnected tweet $t_i$ and a tweet from the FCG for MisT$_j$ (or its prototypical tweet) $t_k$ can be predicted, according to the two options described in Section~\ref{sec:graph_link}. If a link is predicted, it receives the label of MisT$_j$.

We trained the TMD-GLP system to perform graph-link prediction on the training collection of {\sc CoVaxLies} described in Section~\ref{sec:data}. Labeled links within the training data were used as positive examples, and we perform negative link sampling to construct negative examples. Negative link sampling consists of corrupting a link triplet $(t_i,m_j,t_k) \in E$. This process is performed by randomly sampling a different tweet $t_s \in V_{T}$, where $V_{T}$ represents all training tweets, to replace $t_k$. We ensure $(t_i,m_j,t_s) \not\in E$ by re-sampling if we ever sample a link in $E$. This process guarantees that corrupted triplets are not real links. We utilize these negative links for learning, with the goal being to utilize the TMD-GLP system to score labeled links higher than corrupted links.
Moreover, we optimized the following margin loss to train TMD-GLP when  performing graph-link prediction:
\begin{equation} \label{eq:loss}
    \mathcal{L} =\sum\limits_{\substack{(t_i,m_j,t_k) \in E}}{\sum\limits_{\substack{(t_i,m_j,t_s) \not\in E}}\left[\gamma - f(te_i,me_j,te_k) + f(te_i,me_j,te_s)\right]_+}
\end{equation}
where $\gamma$ is a training score threshold which represents the differences between the score of correct links and the incorrect predicted links. The loss $\mathcal{L}$ is minimized with the ADAM\cite{kingma2014adam} optimizer, a variant of gradient descent.

\begin{figure}[t]
    \centering
    \includegraphics[width=0.45\textwidth]{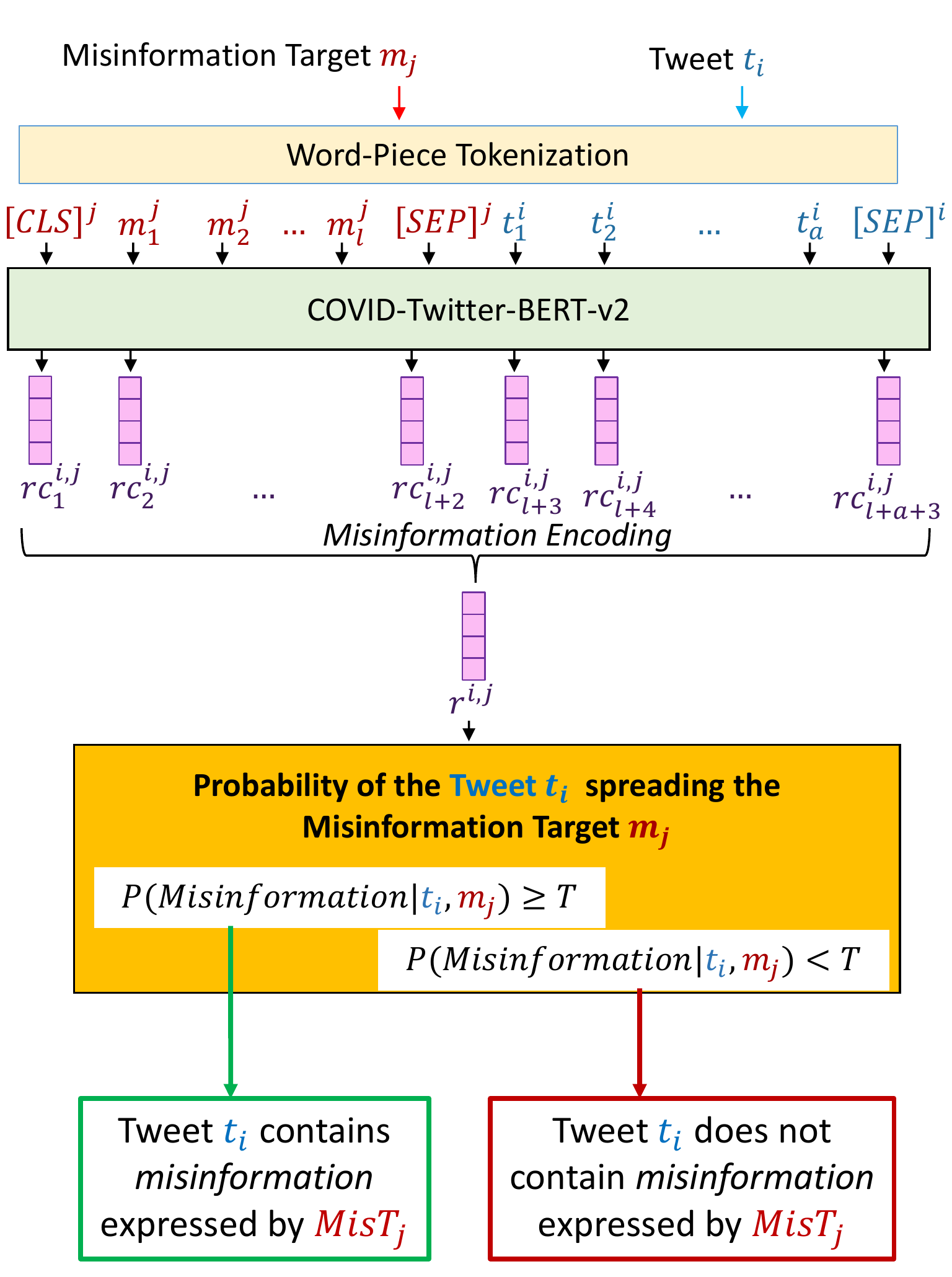}
    \caption{Neural Architecture for Twitter Misinformation Detection as a Binary Classification with BERT (TMD-BC-BERT).}
    \label{fig:baseline}
\end{figure}

\subsection{Misinformation Detection as Neural Classification}
\label{sec:TMD-BC}
Since most existing systems that tackle misinformation detection are binary classifiers, we also generated a baseline binary classification system that can operate on the same data as the TMD-GLP system. 
Therefore we designed a simple neural architecture, 
following prior work \cite{bert-rerank}, 
which directly classifies tweets as containing or not containing the misinformation expressed (explicitly or implicitly) in one of the MisTs encoded in the MKG.  
{This classification uses the text of a tweet along with the description of a MisT to perform sequence-pair classification. }
This simple neural architecture, which we call Twitter Misinformation Detection as Binary Classification with BERT (TMD-BC-BERT), is illustrated in Figure~\ref{fig:baseline}.

Joint Word-Piece Tokenization is performed for both a MisT $m_j$ and a tweet $t_i$. 
This produces a single sequence of word-piece tokens for both the misinformation target and the tweet, {with the misinformation target text and the tweet text separated by a special $[SEP]$ token. The beginning $[CLS]$ token and end $[SEP]$ token are placed at the beginning and end of the joint sequence respectively, as is done in prior work} \cite{bert} {with multiple BERT sequences.}
These tokens are provided to COVID-Twitter-BERT-v2, which produces contextualized embeddings for each word-piece token. 
A single contextualized embedding $r^{i,j}$ represents the entire sequence of tokens from MisT $m_j$ and from $t_i$, which is the first contextualized embedding $rc_1^{i,j}$, {representing the initial $[CLS]^j$ token embedding}.
This embedding is provided to a fully-connected layer with a softmax activation function which outputs a probability distribution over $P(Misinformation|t_i,m_j)$. As Figure~\ref{fig:baseline} shows, misinformation is recognized when the probability is larger than a predefined threshold. 
In our experiments, the value of the threshold $T$ was determined on the development data to be $0.9995$. 
The TMD-BC-BERT system is trained to classify tweets that contain misinformation concerning a given MisT, while using the same training data that was used for training the TMD-GLP system. In addition, the
TMD-BC-BERT system was trained end-to-end using the cross-entropy loss function:
\begin{equation}
    \mathcal{L} = -\sum_{(r, t_i, m_j) \in D}{log{P(r | t_i, m_j)}}
\end{equation}
where $r\in \{Misinformation, NO\_Misinformation\}$ and $D$ is a set of all training tweets judged to contain (or not contain) misinformation addressed by a certain MisT of interest. The loss $\mathcal{L}$ is minimized with ADAM \cite{kingma2014adam}, a variant of gradient descent.

We also compare the TMD-GLP system against a system implemented to  use Long Short-Term Memory (LSTM) \cite{hochreiter1997long} cells instead of COVID-Twitter-BERT-v2 in the architecture illustrated in Figure~\ref{fig:baseline}. 
Specifically, we use $2$ layers of Bi-LSTMs \cite{bilstm} of size $1024$.
We call this baseline system TDM-BC-LSTM.

\section{Experimental Results}
\label{sec:results}

Because we believe that it is critical to detect misinformation about COVID-19 vaccines only in tweets that are truly relevant to the MisTs of interest, we first conducted experiments to evaluate the
quality of retrieval and then we separately evaluated the quality of misinformation detection. 
\subsection{Evaluation of the Tweets Potentially Containing Misinformation about COVID-19 Vaccines}

\begin{table*}[ht]
\centering
\small
\begin{tabular}{l|ccc|c}
    \toprule
    Retrieval System &  Judged Tweet Count &  & & Percent of Tweets  \\
     &  \emph{Relevant} & \emph{Not\_Relevant} & Total & Judged \emph{Relevant}  \\
    \midrule
    BM25 \cite{bm25} & 1,979 & 2,174 & 4,153 & 47.7\%
    \\
    BERTScore (DA) \cite{bertscore} & 1,475 & 3,214 & 4,689 & 31.5\%
    \\
    \hline  
    Total & 3,075 & 4,171 & 7,246 & 42.4\%
    \\
    \bottomrule
\end{tabular}
\caption{Distribution of human judgements of tweet relevance against Misinformation Targets (MisTs).}
\label{tb:ret_results}
\end{table*}

The evaluation of relevant tweets for each MisT from our {\sc CoVaxLies} dataset was performed by considering two retrieval systems: (1) one using the BM25 scoring function, and (2) one using the BERTScore (DA). 
The methods used by these systems was described in Section~\ref{sec:data}. 
In order to conduct the retrieval evaluations, each MisT was used to formulate a query, processed by both retrieval systems. Out of the  753,017 unique tweets from $\cal{C}_{T}$ that we have collected from
the Twitter API, 
the retrieval system using BM25 returned
4,153 tweets it deemed relevant to the 17 MisTs of interest. In contrast, the retrieval system based on BERTScore returned 4,689 tweets that it deemed relevant.
The difference in number of tweets deemed relevant by each system can be explained by how each system treated the replacement of the term ``COVID-19" with ``coronavirus" in the query. The retrieval system using the BM25 scoring function was much less sensitive to the query change, returning only few more  than 200 tweets deemed relevant (on average 244 relevant tweets per MisT), while the system using BERTScore returned slightly more tweets deemed relevant when the query was modified (on average 276 relevant tweets per MisT).

\begin{figure}[t]
    \centering
    \includegraphics[width=0.9\linewidth]{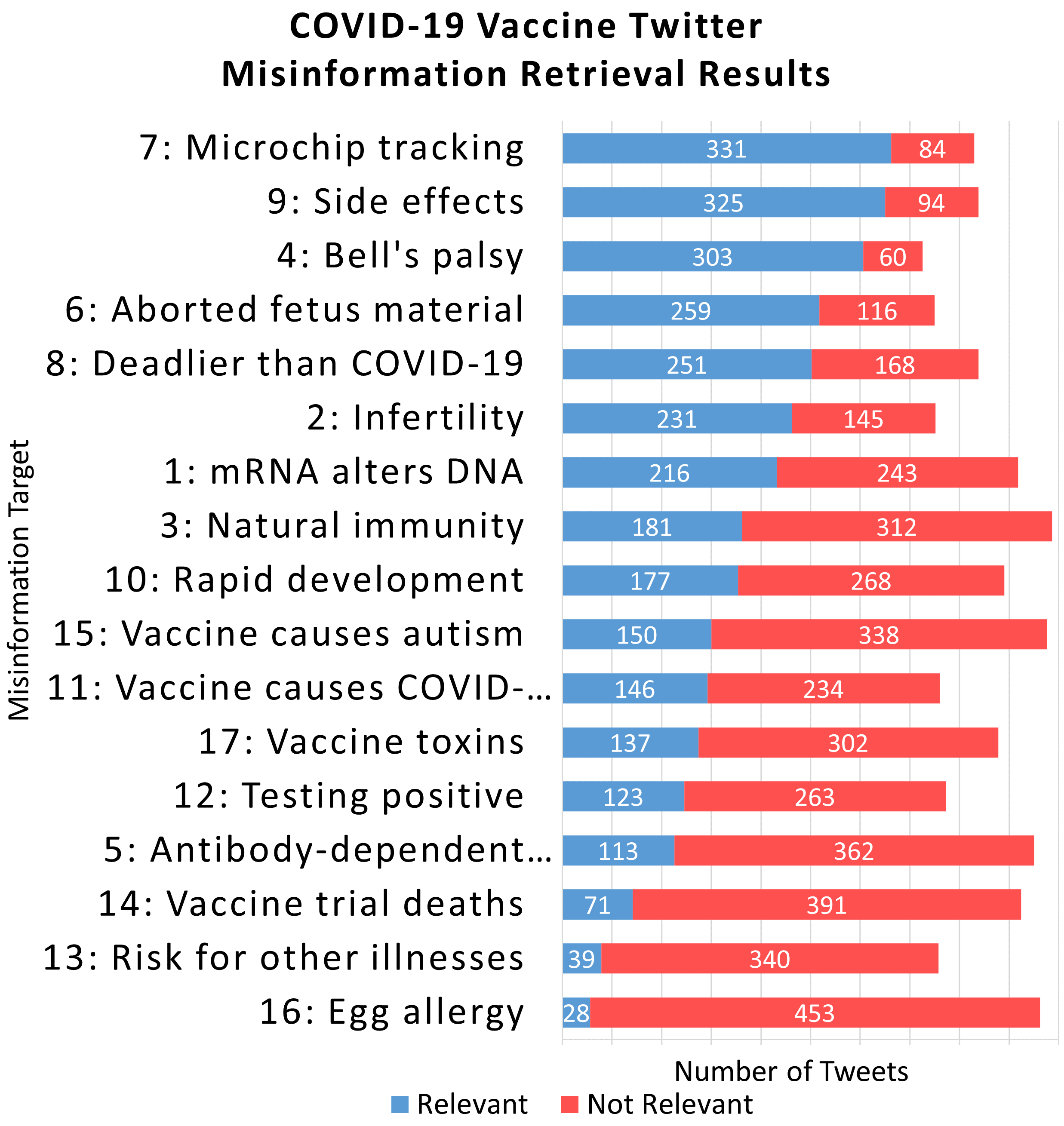}
    \caption{Number of tweets judged relevant or non-relevant by human experts for MisTs of interest. MisTs are presented in descending order of their number of tweets judged relevant.}
    \label{fig:retrieval_targets}
\end{figure}

Human judgements of the relevance of both retrieval systems were performed on the unique tweets 
from the $\cal{T_R}$ collection first mentioned in Section~\ref{sec:data-ir}.   $\cal{T_R}$ contains 7,246 tweets {\em deemed}  relevant to our MisTs of interest by at least one of the retrieval systems. 
Each human judgement has established the relevance or non-relevance of each tweet against the MisTs of interest. Three natural language experts participated in the judgements of relevance. 
Their inter-annotator agreement was discussed in Section~\ref{sec:data-ir}.
Table~\ref{tb:ret_results} lists the judgement results on the $\cal{T_R}$ collection.
We see that the human judges have found a similar number of MisT-\emph{Relevant} tweets from the results returned by the retrieval system using the BM25 scoring function when compared to the retrieval system using BERTScore (DA) (1,979 vs 1,475), while the number of retrieved tweets judged non-relevant by the system using BERTScore (DA) is significantly larger than the number corresponding to the tweets judged as  \emph{Not\_Relevant} from the tweets retrieved by the system using the BM25 scoring function (3,214 vs 2,174).
This indicates, as shown in Table~\ref{tb:ret_results} that the percent of tweets deemed relevant by the system using the BM25 scoring function and also judged relevant by human experts is much higher than the same percent of tweets deemed relevant by the system using BERTScore. Therefore, retrieving tweets with the system that uses the BM25 scoring function is far better than using a retrieval system informed by BERTScore. 

To provide additional details, Figure~\ref{fig:retrieval_targets} illustrates the total number of tweets judged \emph{Relevant} and \emph{Not\_Relevant} for each MisT. As shown in the Figure, for some MisTs, such as MisT$_7$,  
MisT$_9$, or MisT$_4$, the majority of the tweets retrieved using both retrieval systems were judged relevant. In contrast, for other MisTs, the results of the retrieval systems were judged mostly \emph{Not\_Relevant}, e.g. for MisT$_5$, MisT$_{14}$, MisT$_{13}$, or MisT$_{16}$. 

\subsection{Evaluation of Detecting Misinformation about COVID-19 Vaccines}
\label{sec:eval-misinfo}

Evaluation was performed by considering that misinformation detection was cast as multi-label binary classification. 
This means that multiple MisTs could be predicted for every tweet $t_i$ from the test collection of {\sc CoVaxLies}, some correct and others not. 
Predictions of graph links of a tweet $t_i$ to various MisTs were compared against the MisTs with which the
tweet $t_i$ was paired in the test collection of {\sc CoVaxLies}. 
For example, if the test collection contained only one pair in which $t_i$ was associated with one MisT, e.g. [$t_i$, MisT$_a$], then during evaluation, we compared MisT$_a$ against the MisTs predicted for $t_i$. 
If MisT$_a$ was predicted, then that prediction was correct and counted as a true positive ($tp$). 
If MisT$_a$ was not predicted, then that prediction was incorrect and counted as a false negative ($fn$).
If other MisTs were predicted which were different from MisT$_a$, then each of those predictions were incorrect and counted as a false positive ($fp$).

In a similar way we have evaluated the prediction of graph links for tweets that were paired with multiple MisTs in the testing data.
For example, if the test collection contains a tweet $t_i$ paired with $k$ MisTs, e.g. [$t_i$, MisT$_1$], [$t_i$, MisT$_2$], ..., [$t_i$, MisT$_k$], then during evaluation, we compared these MisTs against the MisTs predicted for $t_i$. 
Each predicted MisT which coincides with one of the $k$ MisTs paired with $t_i$ in the test collection is counted as a true positive.
If any of the $k$ MisTs paired with $t_i$ was not predicted,  then it is counted as a false negative.
Any predicted MisT which was different from the $k$ MisTs paired with $t_i$ in the test collection  counted as a false positive.
The evaluation metrics of Precision and Recall using the true positives, false positives, and false negatives were computed as:
\begin{equation}  \label{eq:p_r}
    Precision=\frac{tp}{tp + fp}; Recall=\frac{tp}{tp + fn}; 
\end{equation}
Because the number of paired MisTs varies across tweets in the testing collection, this learning task could not be cast as a multi-class classification problem, but rather as a multi-label binary classification problem.
System performance was evaluated using Micro averaged Precision (P), Recall (R) and F$_1$
\footnote{F$_1$ is defined as $F_1 =2 \times P \times R/(P+R)$} score. The evaluation results are provided in Table~\ref{tb:results}. 

When evaluating the Twitter Misinformation Detection as Graph Link Prediction (TMD-GLP) system, we have considered (a) four possible knowledge embedding models (TransE; TransD; TuckER; and TransMS), which provide different graph-link scoring functions; and (b) two possible options of selecting a tweet from a FCG corresponding to a MisT to which an unconnected tweet could be linked. In option 1: we select {\em all} tweets connected to the FCG, while in {option 2}: we select a {\em prototypical} tweet of the FCG. 
We also evaluate the KNN baseline configuration of the TM-GLP system, which does not consider edge types between connected tweets in the \emph{Misinformation Knowledge Graph}.
Hence, we have evaluated nine different configurations of the TMD-GLP system. 
In addition, we have evaluated the system for Twitter Misinformation Detection as Binary Classification, with both BERT (TM-BC-BERT) and LSTM cells (TMD-BC-LSTM). The evaluation results of all these configurations of the 
TMD-GLP system are listed in Table~\ref{tb:results} along with the results of the TMD-BC-BERT and TM-BC-LSTM systems. The bolded numbers represent the best results obtained across all systems.

\begin{table}[t]
\centering
\small
\begin{tabular}{l|ccc}
    \toprule
    Misinformation Detection & Micro & & \\
    System &  Precision &  Recall & F$_1$-Score  \\
    \midrule
    BM25-BC \cite{covidlies} & 42.9 & 63.7 & 51.2
    \\
    BERTScore(DA)-BC \cite{covidlies} & 25.9 & 32.9 & 29.0
    \\
    \hline  
    TMD-BC-LSTM & 47.6	& 64.8	& 54.9
    \\
    TMD-BC-BERT & 68.9	& 86.7	& 76.8
    \\
    \hline  
    TMD-GLP & 	& 	& 
    \\
     + KNN & 80.2 & 82.2 & 81.2
    \\
     + TransE-All & 77.8 & 87.5 & 82.4
    \\
     + TransE-Prototypical & 77.0 & 85.7 & 81.1
    \\ 
     + TransD-All & 71.9 & 84.8 & 77.8
    \\ 
     + TransD-Prototypical & 68.8 & \textbf{88.3} & 77.4
    \\ 
     + TuckER-All & 64.0 & 83.8 & 72.6
    \\ 
     + TuckER-Prototypical & 61.5 & 85.4 & 71.5
    \\ 
     + TransMS-All & \textbf{82.9} & 85.4 & 84.1
    \\ 
     + TransMS-Prototypical & 82.4 & 86.4 & \textbf{84.3}
    \\ 

    \bottomrule
\end{tabular}
\caption{Evaluation results for misinformation detection experiments performed on the test collection from the {\sc CoVaxLies} dataset described in Section~\ref{sec:data}.}
\label{tb:results}
\end{table}


As shown in Table~\ref{tb:results}, the TMD-BC-LSTM and TMD-BC-BERT systems had higher Precision, Recall, F$_1$ scores than the BM25-BC and BERTScore(DA)-BC systems, demonstrating the importance of fine-tuning neural misinformation detection systems. 
The TMD-BC-BERT system significantly outperformed the TMD-BC-LSTM system, yielding a Micro F$_1$ score of $76.8$, which is superior to the Micro F$_1$ score of the TMD-BC-LSTM system of $54.9$, but the F$_1$ results of the TMD-BC-BERT system were lower than 7 out of 9 configurations of the TMD-GLP system. 
The only configurations of the TMD-GLP system under-performing the TMD-BC-BERT system used TuckER as the knowledge embedding model. 
The TMD-GLP with the KNN baseline configuration outperformed TMD-BC-BERT, with a Micro F$_1$ score of $81.2$, which also performed better than 5 out of the 8 TMD-GLP configurations which considered MisTs as edge types in the \emph{Misinformation Knowledge Graph}. 
From the evaluation results obtained for the TMD-GLP system we can 
see that when selecting the prototypical tweet of the FCG of a MisT, the results were inferior to those obtained when selecting any tweet from the FCG, with the exception of the system configuration using the TransMS knowledge embedding model. In fact, the best overall results for misinformation detection were obtained when graph link prediction was used, informed by the scoring function of the TransMS knowledge embedding system, and when selecting the prototypical tweet of each MisT-specific FCG. 
For this configuration, the TMD-GLP system produced a Micro F$_1$ score of $84.3$. This can be explained because the TransMS knowledge embedding model accounts for the most interactions between nodes and edges in its scoring function, seen in Equation~\ref{eq:transms}, while also best modeling symmetric relationships, which we often see in our \emph{Misinformation Knowledge Graph} (MKG). This may also explain why the TransMS-All configuration of the TMD-GLP system generated the best overall Precision score.

Additionally, two baselines for Detecting Misinformation about COVID-19 Vaccines were considered, both assuming that relevance retrieval is sufficient. 
These systems follow prior work \cite{covidlies} and cast misinformation detection as misinformation retrieval, by considering each tweet as a query and returning the most relevant MisTs. We 
acknowledge that this retrieval framework is atypical because the collection of MisTs is several orders of magnitude smaller than the number of tweets from the test collection of {\sc CoVaxLies}, whereas retrieval systems typically work on an index which is much larger than the number of queries. But since this is the framework for retrieval that was considered in \cite{covidlies}, we adopted the same
framework for these baselines. 
In our experiments, the retrieval system using the BM25-BC model returned for each tweet  the most relevant MisTs, and we considered that MisTs with a relevance score above a pre-defined threshold $T$ are predicted as linked to the tweet.
The BERTScore(DA)-BC baseline compares the text of each tweet against the text of each MisT using the BERTScore (DA) relevance model. BERTScore (DA) assigns a relevance score to each MisT for each tweet, and MisTs with a relevance score above a pre-defined threshold $T$ are predicted as relevant.
The threshold $T$ for both systems is selected by maximizing F$_1$ score on the development set. 

For this purpose, we have evaluated on the test collection of {\sc CoVaxLies} the retrieval systems using the (1) BM25 scoring function, used for Binary Classification (BM25-BC) or; (2) the BERTScore(DA) scoring function, used for Binary Classification (BERTScore(DA)-BC). 
The retrieval system using the BM25 score produced a Micro F$_1$ score of $51.2$, setting a baseline expectation of performance. This performance can be attributed to the large amount of shared terminology between MisTs and \emph{Relevant} tweets, which benefits the BM25 scoring function. 
The retrieval system using BERTScore (DA) produced a Micro F$_1$ score of $29.0$, which was much lower than expected, when comparing with results published in prior work \cite{covidlies}.

The results of the evaluations listed in Table~\ref{tb:results} are interesting, as they generally indicate that casting misinformation detection as graph-link prediction, informed by knowledge embedding models such as TransE, TransD, and TransMS, can generate promising results, superior to the results obtained when considering misinformation detection as a multi-label binary classification problem, as most current systems do. We can also notice 
that TransE, a much simpler knowledge embedding model, outperformed TransD and TuckER, and was competitive with TransMS in their impact on misinformation detection. This also indicates that some advancements in knowledge embedding models do not necessarily result in improved performance on certain tasks such as misinformation detection.

System hyper-parameters were selected by maximizing the F$_1$ score of each system on the development collection. 
The TMD-BC-BERT and TMD-GLP systems share many of the same training hyper-parameters: a linearly decayed learning rate of $5e-4$ which was warmed up over the first $10\%$ of the $40$ total epochs, an attention drop-out rate of $10\%$, a batch size of $6$, and a gradient norm clipping of $1.0$. 
The TMD-BC-LSTM system has $2$ layers of bidirectional LSTM cells with a size of $1024$, a linearly decayed learning rate of $5e-2$ which was warmed up over the first $10\%$ of the $40$ total epochs, a batch size of $6$, and a gradient norm clipping of $1.0$.
The TMD-GLP system has different knowledge embedding hyper-parameters depending on the knowledge embedding model selected.
The tweet knowledge embedding size $z$ was set to $8$ for all knowledge embedding models except for TransD, which used $16$ such that both the split knowledge embeddings have size $8$. 
The MisT knowledge embedding size $v$ was set to $8$ for all knowledge embedding models except for TransD, which used a knowledge embedding size of $16$ for the same reason as above, and TransMS, which used a knowledge embedding size of $9$ to ensure the MisT knowledge embedding vectors were size $8$ when considering the extra $\alpha_j$ parameter. 
Initial experiments on the development collection of {\sc CoVaxLies} produced the best results for embedding size of 8. We experimented with embedding sizes of 4, 8, 16, 32, 64, and 128, and found that performance was, on average, 2-4 F$_1$ points lower on the development set as the size increased above 8 when using the TransMS-Prototypical configuration. We observed the same decrease in results for the TransE-All configuration. Embeddings of size 4 produced a 6-point F$_1$ score drop on the development set for the TransMS-Prototypical configuration. Therefore we decided that we would use embedding sizes of 8 for all test collection experiments. 
Our intuition is that the relatively simple FCGs of the misinformation knowledge graph likely do not need larger embeddings to encode the necessary graph link prediction, but cannot encode that information in embeddings of size 4.

The TMD-GLP system utilizes the training collection for learning to perform graph-link prediction by optimizing the margin loss, described in Equation~\ref{eq:loss}.
The $\gamma$ hyper-parameter is set to $1.0$ for all knowledge graph embedding models, and we sample $1$ negative corrupted link for each labeled link.
Threshold values $N_x$ and $T_x$ were also automatically selected by maximizing the F$_1$ score of the TMD-GLP system on each MisT$_x$ on the development collection.

\subsection{Discussion of the Identification of Tweets Potentially Containing Misinformation about COVID-19 Vaccines}
Collecting tweets for {\sc CoVaxLies} and judging their relevance against each MisT revealed a large discrepancy between retrieval performance of the retrieval systems using the BM25 scoring function or BERTScore. 
Prior work \cite{covidlies} on the {\sc COVIDLies} dataset found that retrieval using BERTScore performed better than retrieval using the BM25 scoring function for misinformation detection, but the entire {\sc COVIDLies} dataset was collected using only BERTScore. 
Their collection methodology resulted in a judged \emph{Relevant} tweet percentage of $14.98\%$, meaning only $14.98\%$ of their discovered tweets were judged \emph{Relevant} by annotators.
We found a similar \emph{Relevant} tweet percentage of $31.5\%$ on our collection when only retrieving tweets using the BERTScore (DA), but simultaneously identified that the retrieval system using the BM25 scoring function generated a relevant percentage of $47.7\%$, presented in Table~\ref{tb:ret_results}.
This large difference indicates that there are many instances of tweets containing misinformation which are not discovered by the retrieval system using BERTScore, and that the retrieval system using the BM25 scoring function is actually better at discovering more \emph{Relevant} tweets for each MisT. 
We also analyzed the differences in total tweets: the retrieval system using the BM25 scoring function returned a total of $4,153$ tweets, while the retrieval system using BERTScore (DA) returned $4,689$ tweets. 
This difference in the number of retrieved tweets arises from the sensitivity of each system to changes to the query: We found that the retrieval system using BERTScore to be more sensitive to replacing ``COVID-19" with ``coronavirus" when querying the {\sc CoVaxLies} dataset for each MisT. The retrieval system using BERTScore produced more disjoint retrieved lists of tweets for the original and the modified query, while when the retrieval system using the BM25 scoring function was used, there was much more overlap. But, when using the retrieval system informed by BERTScore,  we discovered that most of these additional non-overlapping tweets were judged as \emph{Not\_Relevant}.

Advantages remain to using BERTScore, as the \emph{Relevant} tweets it finds are characterized by less term overlap with the language used to describe the MisTs, and, thus it emphasizes more semantic relevancy. However, ignoring classical scoring functions such as BM25 leads to a heavily biased dataset of tweets that potentially contain misinformation, which may very well lead to misinformation topic shift, such that the tweets are deemed relevant, when in fact they share more semantics with other topics than the one we try to collect, namely misinformation about COVID-19 vaccines.  
As provided in Table~\ref{tb:results}, performance of the misinformation detection systems using the BM25 scoring function and BERTScore (DA) on misinformation detection on our {\sc CoVaxLies} dataset was significantly different. 
The BM25-BC system scored much higher on all metrics than BERTScore(DA)-BC, which was the opposite conclusion drawn from prior work \cite{covidlies} on the {\sc COVIDLies} dataset. 
We hypothesized that this difference was due to the data collection methodology utilized to create the {\sc COVIDLies} dataset: 
The only retrieval system used to find tweets for each MisT was BERTScore, therefore the \emph{Relevant} annotations would be biased towards a BERTScore-based model. 
This would naturally lead the misinformation detection system using the BM25 scoring function to perform worse than a BERTScore-based model in the evaluation of misinformation detection.
To test this hypothesis, we modified our data collection to only include annotated tweets which were discovered by BERTScore (DA) during the retrieval of relevant tweets for each MisT. 
We re-ran both the BERTScore(DA)-BC and BM25-BC misinformation detection systems on this modified collection of {\sc CoVaxLies} and report the results in Table~\ref{tb:results_bertscore}. 
BM25-BC produces a F$_1$ score of $32.0$, while BERTScore(DA)-BC produces an F$_1$ score of $39.2$. 
This experiment demonstrates the flaw of collecting data using only BERTScore (DA): The collection will be heavily biased towards \emph{Relevant} tweets which BERTScore (DA) alone would discover. Simultaneously, \emph{Relevant} tweets would be missed by BERTScore but would have been discovered by other systems such as BM25. This phenomenon can lead to misleading conclusions during system evaluation.

\section{Discussion}
\label{sec:dis}

\begin{table}[t]
\centering
\small
\begin{tabular}{l|ccc}
    \toprule
    Misinformation & Micro & & \\
    Detection System &  Precision &  Recall & F$_1$-Score  \\
    \midrule
    BM25-BC & 19.7 & \textbf{85.5} & 32.0
    \\
    BERTScore(DA)-BC & \textbf{29.9} & 57.1 & \textbf{39.2}
    \\
    \bottomrule
\end{tabular}
\caption{Evaluation results of misinformation detection on tweets from the {\sc CoVaxLies} dataset discovered by BERTScore(DA).}
\label{tb:results_bertscore}
\end{table}

\subsection{Assessing the Quality of Misinformation Detection}
The experimental results, provided in Table~\ref{tb:results}, indicate that detecting misinformation using the TMD-GLP system, that benefits from graph link prediction,  improves upon baseline systems using only retrieval scoring or neural classification approaches. 
The TransE knowledge embedding model had a better impact than the TransD or TuckER knowledge embedding models on the TMD-GLP system, although it is a much older and simpler model. 
We believe this is due to the fact that knowledge embedding models have been rarely tested in extrinsic evaluations, in which they are used for other applications than knowledge representation per se. 
Our goal was not to model a massive, multi-relational knowledge graph with complicated hierarchies and directed edges. 
Our Misinformation Knowledge Graph (MKG) is relatively simple, containing a Fully Connected Graph for each MisT, and therefore benefiting from a simpler knowledge embedding model, such as TransE. 
We also see this pattern hold for the KNN baseline configuration, where we entirely ignore the MisT edge embeddings in our MKG. 
But, TransMS, an improved knowledge graph embedding model, did improve upon TransE and KNN in our misinformation detection framework.  
We believe TransMS performed best for two reasons:

\begin{itemize}
\item The scoring function for TransMS, provided in  Equation~\ref{eq:transms}, is insensitive to the direction of the relationship. Symmetric relationships, which are the only kind of relationships in our MKG, benefit from this property.

\item Non-linear interactions in the scoring function for TransMS, in the form of the $tanh$ operations, provide more rich knowledge embeddings than the linear, multiplicative, and dynamic projection approaches of TransE, TuckER, and TransD respectively. 
\end{itemize}

\begin{table}[t]
\centering
\small
\begin{tabular}{l|ccc}
    \toprule
    Misinformation & Micro & & \\
    Detection System &  P &  R & F$_1$  \\
    \midrule
    BM25-BC & 15.8 & 24.1 & 19.1
    \\
    BERTScore(DA)-BC & 13.3 & 22.2 & 16.6
    \\
    \hline  
    TMD-BC-BERT & 21.8 & 54.1 & 31.1
    \\
    \hline  
    TMD-GLP &  &  & 
    \\
    + TransMS-Prototypical 	&	\bf{29.4} & \bf{79.7} & \bf{40.7} 
    \\
    \bottomrule
\end{tabular}
\caption{Results from 5-fold cross validation  on the COVIDLies dataset for misinformation detection.}
\label{tb:results_covidlies}
\end{table}


\begin{table}[t]
\centering
\small
\begin{tabular}{l|cccc}
    \toprule
    ID &  Precision &  Recall & F$_1$-Score & n$_x$  \\
    \midrule
    1	&	87.9	&	76.3	&	81.7	&	13 \\
    \hline  
    2	&	92.9	&	100.0	&	96.3	&	20 \\
    \hline  
    3	&	76.9	&	87.0	&	81.6	&	26 \\
    \hline  
    4	&	94.5	&	100.0	&	97.2	&	24 \\
    \hline  
    5	&	79.3	&	95.8	&	86.8	&	6 \\
    \hline  
    6	&	100.0	&	100.0	&	100.0	&	24 \\
    \hline  
    7	&	97.1	&	84.6	&	90.4	&	20 \\
    \hline  
    8	&	85.3	&	55.8	&	67.4	&	19 \\
    \hline  
    9	&	63.4	&	88.1	&	73.8	&	27 \\
    \hline  
    10	&	71.8	&	80.0	&	75.7	&	15 \\
    \hline  
    11	&	80.6	&	86.2	&	83.3	&	14 \\
    \hline  
    12	&	81.8	&	75.0	&	78.3	&	12 \\
    \hline  
    13	&	0.0	    &	0.0	    &	0.0	    &	2 \\
    \hline  
    14	&	43.5	&	90.9	&	58.8	&	4 \\
    \hline  
    15	&	91.4	&	97.0	&	94.1	&	7 \\
    \hline  
    16	&	44.4	&	100.0	&	61.5	&	1 \\
    \hline  
    17	&	66.7	&	76.9	&	71.4	&	11 \\
    \bottomrule
\end{tabular}
\caption{Misinformation detection performance of TMD-GLP with the TransMS-Prototypical configuration for each MisT in the test collection. n$_x$ is the number of nodes in the FCG for MisT$_x$}
\label{tb:detailed_results}
\end{table}

\begin{table*}[t]
\centering
\small
\begin{tabular}{p{0.2\linewidth} | p{0.4\linewidth} | c|c|c}
    \toprule
    {\bf Judged MisT} & {\bf Tweet} & {\bf BM25-BC} & {\bf TMD-BC-BERT} & {\bf TMD-GLP} \\
    \hline
    1: RNA alters a person's DNA when taking the COVID-19 vaccine.
    & 
    The Covid vaccine is not a vaccine.
    It's gene therapy making genetically modified humans
    & 
    \xmark & \cmark & \cmark \\
    \hline
    17: Vaccines contain unsafe toxins such as formaldehyde, mercury or aluminum.
    & 
    Covid vaccine: think 5x over currently.  You cannot detox from this.  You will be inserted with nano technology. Nano lipids. You'll become a human antenna with the aluminum encased in the nano lipids.  They are basically impossible to remove once they are in.
    & 
    \xmark & \xmark & \cmark \\
    
    \hline
    13: The COVID-19 vaccine can increase risk for other illnesses.
    & 
    The covid vaccine is a one way ticket to cancer and dementia. Remember that. Don't vaccinate.
    & 
    \xmark & \xmark & \xmark \\
    
    \bottomrule
\end{tabular}
\caption{Tweets in which MisTs are not correctly detected by BM25-BC, TMD-BC-BERT, and TMD-GLP with the TransMS-Prototypical configuration.}
\label{tb:errors}
\end{table*}

Additionally, we observed that TransMS was the only knowledge embedding model which performed better when selecting a prototypical tweet from a MisT-informed FCG (as opposed to considering all the tweets of the FCG) to predict a link to an unconnected tweet. 
This can be explained by the observation that a prototypical tweet is represented as the average knowledge embedding of all tweets from the same MisT-informed FCG, which can create a more ``coherent" representation of the MisT, but may damage some of the specific structure of the knowledge embeddings for each tweet in the MisT-informed FCG needed for link prediction.
In contrast, when all the tweets of the MisT-informed FCG are considered, some of their knowledge embeddings may be less of a ``coherent" representation of the MisT, but the stronger specific knowledge embedding structure remains for link prediction. 
Thus, the Prototypical approach likely reduces performance when the gains in ``coherence" are outweighed by losses to knowledge embedding structure, as we see when using the TransE, TransD, or TuckER knowledge embedding models. 

\subsection{Portability of Misinformation Detection as Graph Link Prediction to Other Datasets}

In order to assess the portability of our approach to detecting misinformation as graph link detection to other datasets, we have considered the {\sc COVIDLies} dataset \cite{covidlies}.  The {\sc COVIDLies} collection consists of 5,748 tweets annotated with stance towards COVID-19 pandemic misinformation targets, with stance values of \emph{``Agree"}, \emph{``Disagree"}, and \emph{``No Stance"}.
However, the {\sc COVIDLies} annotators made no distinction between tweets which contained a neutral \emph{``No Stance"} towards a MisT and tweets which were not relevant to a MisT, which were also annotated as \emph{``No Stance"}. To resolve this problem, we categorized tweets labeled as \emph{``Agree"} or \emph{``Disagree"} with a MisT as \emph{``Relevant"} to the MisT, while all tweets with \emph{``No Stance"} against a Mist were considered \emph{``Not Relevant"} to that MisT.
This re-annotation of {\sc COVIDLies} in terms of relevance of tweets towards various MisTs led to 
having only 17\% of the unique tweets from {\sc COVIDLies} as relevant to  one or more MisTs. 
In comparison, in the {\sc CoVaxLies} collection, 57\% of the unique tweets are relevant to one or more MisTs. 

The portability of the system for Twitter Misinformation Detection as Graph Link Prediction (TMD-GLP) 
on the {\sc COVIDLies} collection also entails the availability of training, development and test sets 
on this collection. However, {\sc COVIDLies} has no official training or testing collections. Therefore,
we split {\sc COVIDLies} into 5 evenly distributed folds, assigning three folds for training, one fold
for development and the fifth fold for testing. This process was performed five times, enabling the evaluation of the results of TMD-GLP on {\sc COVIDLies}. Results of the evaluations, when considering the configuration of the TMD-GLP that produced the best results on the {\sc CoVaxLies} dataset, namely when using the TransMS knowledge embedding model with a prototypical representation for the FCG of each MisT, i.e. TMD-GLP+TransMS-Prototypical, are listed
in Table~\ref{tb:results_covidlies}.  We also ported on {\sc COVIDLies} the system for 
Twitter Misinformation Detection as Binary Classification with BERT (TMD-BC-BERT), evaluating it on the
same 5-fold cross validation as we did with the TMD-GLP+TransMS-Prototypical system. 

Table~\ref{tb:results_covidlies} shows that even on {\sc COVIDLies}, the TMD-GLP+TransMS-Prototypical system performs best, obtaining an F$_1$ score of $40.7$. However,
a major performance drop is observed between the performance of this system from its  operation on the {\sc CoVaxLies} dataset to its operation on the {\sc COVIDLies} dataset, namely a drop from an F$_1$ score of $84.3$ to an F1 score of $40.7$. 
This drop can be explained when the dataset statistics are considered. Because the {\sc COVIDLies} collection has a smaller percentage of relevant tweets than the {\sc CoVaxLies} collection, it will lead to the discovery of a  \emph{Misinformation Knowledge Graph} containing smaller FCGs for each MisT than those discovered when using the {\sc CoVaxLies} dataset. This entails that it becomes harder to predict a correct graph link, when the FCGs are smaller.  
We can notice that both TMD-BC-BERT and TMD-GLP systems suffer from this major reduction in relevant tweets in the {\sc CoVaxLies} collection, as the results listed in Table~\ref{tb:results_covidlies} show that both systems generate much better Recall results than Precision results, which are quite low. 
This low Precision is likely due to such a small number of relevant tweets for each MisT in {\sc COVIDLies}, with many MisTs having only one or two relevant tweets from which to learn to predict true positive links, leading to many more false positive links.
But, the evaluation results listed in Table~\ref{tb:results_covidlies} showcase the portability of the TM-GLP method on a second dataset, while also highlighting the limitations of the {\sc COVIDLies} dataset.

\subsection{Discussion of Detailed Results and Error Analysis}
Detailed performance of the TMD-GLP system using the TransMS-Prototypical configuration is provided for each MisT in Table~\ref{tb:detailed_results}. We also include the size $n_x$ of the FCG for each MisT.
The TMD-GLP+TransMS-Prototypical misinformation detection system performed best on MisT$_2$: { ``The COVID-19 vaccine causes infertility or miscarriages in women."};  MisT$_4$: ``The COVID-19 vaccine causes Bell’s palsy.", and MisT$_6$: ``The COVID-19 vaccine contains tissue from aborted fetuses.".  
These MisTs made easily identifiable claims, such as ``causing infertility", ``causing Bell's palsy", and ``containing aborted fetus tissue". 
Statements supporting, refuting, or reporting on these claims were very easy to detect. 
For example, the following is a tweet referring to MisT$_2$: {\em ``@JoPatWar @allen40\_allen @Telegraph The Pfizer CEO said that the coronavirus vaccine can make women infertile.
The COVID-19 vaccine} \underline{\em {causes infertility or miscarriages in women." }}, where the claim articulated in MisT$_2$ is underlined.
MisT$_2$, MisT$_4$, and MisT$_6$ also all inform relatively large FCGs, with $n_x \geq 20$, providing a significant sample of tweets to compute prototypical embeddings. 
Additionally, Figure~\ref{fig:retrieval_targets} demonstrates that MisT$_2$, MisT$_4$, and MisT$_6$ are among the upper half of MisTs which had the most \emph{Relevant} tweets judged, meaning they had more relevant data during training than other MisTs.  

The TMD-GLP+TransMS-Prototypical misinformation detection system performed worst on MisT$_{13}$: ``The COVID-19 vaccine can increase risk for other illnesses."; MisT$_{14}$: ``Many people already have died from the COVID-19 vaccine trials."; and MisT$_{16}$: ``The COVID-19 vaccine should not be taken by people who are allergic to eggs.". 
Misinformation claims made by these MisTs were much more difficult to identify because the articulation of the claims was either too vague or too specific. 
MisT$_{13}$ contains the vague statement ``can increase the risk for other illnesses", which was much more vague than the claims made in other MisTs, such as MisT$_4$: ``The COVID-19 vaccine causes Bell’s palsy." 
In contrast, MisT$_{16}$ focuses on specifically ``egg allergies", which was difficult to find on Twitter, as most discussions surrounded other specific allergies. 
MisT$_{13}$, MisT$_{14}$, and MisT$_{16}$ also all inform relatively small FCGs, with $n_x \leq 4$, providing too small a sample of tweets to compute prototypical embeddings. 
Additionally, Figure~\ref{fig:retrieval_targets} demonstrates that MisT$_{13}$, MisT$_{14}$, and MisT$_{16}$ are the three MisTs with the fewest \emph{Relevant} tweets judged, meaning they had less relevant data during training than other MisTs.

Table~\ref{tb:errors} lists tweets which were judged to refer to some  MisT of interest, but the misinformation detection systems that we evaluated failed to identify. 
The first tweet listed in Table~\ref{tb:errors} was judged to refer to MisT$_1$ (also listed in the Table). The TMD-BC-BERT system as well as the TMD-GLP system with the TransMS-Prototypical configuration were able to detect the connection to this MisT, while the baseline using the BM25 scoring function failed to accomplish this task. A lack of exact term overlap between the tweet text and the MisT text explains why a term-based Lucene index, utilized by the BM25 system, along with a BM25 scoring function would be unlikely to discover this misinformation. 
The second tweet from Table~\ref{tb:errors} was judged to refer to MisT$_{17}$. The TMD-GLP system with the TransMS-Prototypical configuration was the only system able to identify the reference to this MisT. The TMD-GLP system compared the knowledge graph embedding of the tweet with the knowledge embedding obtained for other tweets connected to the same MisT-informed FCG, such as {\em ``@FayCortez @annaedney @LauermanJohn @business Because the people who created the deadly, depopulating Covid vaccine are part of the same contingent of Planners who have been spraying you with aluminum, barium, and strontium via chemtrails to intentionally increase your chances of getting Alzheimer's." } The 
TMD-GLP system was able to identify that the second tweet listed in Table~\ref{tb:errors} likely refers to the same MisT as this tweet, and therefore correctly detected the misinformation. 
The third tweet listed in Table~\ref{tb:errors} was judged to refer to MisT$_{13}$. 
None of the systems that we evaluated were able to identify that this tweet referred to MisT$_{13}$. 
There is little term overlap or contextual clues to indicate that this tweet would be related to MisT$_{13}$, which explains why the TMD-BC-BERT systems as well as the system using the BM25 scoring function failed. To determine why the TMD-GLP system failed, we can look at the FCG informed by MisT$_{13}$ and recognize the following: Table~\ref{tb:detailed_results} states that there were only two tweets in the FGC for MisT$_{13}$, and upon further inspection we see that these two tweets only mention the COVID-19 vaccine increasing the risk of ``cancer", ``heart disease", and ``HIV". The third tweet discusses ``cancer", but it also proposes the COVID-19 vaccine is ``a one way ticket to" the illness ``dementia". The fact that ``cancer" or ``dementia" were not recognized as instances of ``illness", because no clinical language processing was applied on the content of the tweets or of the description of the MisT$_{13}$ explains why the third tweet was not linked to MisT$_{13}$ by any of the systems that we have evaluated. 

\subsection{Limitations}

There are several important limitations to our study. The first limitation originates in the fact that
we aim to discover tweets that discuss or refer only to {\em known} misinformation targets.  When additional misinformation targets become known, new relevant tweets must be retrieved, judged if they discuss information that is relevant to the new targeted misinformation, or not, and then enable the creation of new training, development and testing data for 
identifying additional tweets that discuss the new targeted misinformation. The recognition of new,
yet unknown misinformation is not withing the scope of this study. Even if this may seem a major limitation, it is a significant departure from most previous methods, discussed in Section~\ref{sec:related-work}, that detect only {\em if} there is some misinformation (or rumor) in a tweet, but fail to recognize what {\em kind} of misinformation is discussed or referred.

An additional limitation can be found in the use of specific search keywords, such as ``covid" or ``coronavirus". More search terms, such as ``corona", ``vax", or ``jab" may be used more often by the general public, and should be considered in future studies. These terms might reveal new misinformation, or may be used more often by users which tweet about known misinformation.

Another limitation of the study is determined by the fact that we decided not to consider the stance of
the tweets against the misinformation targets. We only recognize if the information shared by a tweet is relevant to a MisT, but do not recognize if it agrees or disagrees with the predication of the MisT, or if it has no stance at all. We believe that stance detection is a separate task, which can be performed only on the tweets that are known to be relevant to a MisT. Previous work \cite{covidlies}, \cite{our-stance} showed that identifying the tweet stance towards a MisT benefits from knowing that the tweet is discussing information relevant to the MisT.
In future work we plan to address the problem of recognizing tweets that are relevant to {\em new} misinformation targets, casting the problem as a zero-shot learning problem. 

\section{Conclusion}

It has been estimated that the COVID-19 vaccines will need to be
accepted by at least 55\% of the population to provide herd immunity, with estimates reaching as high as 85\% depending on country
and infection rate \cite{kwok}.
Reaching these required vaccination levels is hindered by 
vaccine hesitancy across the world \cite{lancet}, which is often fuelled by 
misinformation spreading on social media. Therefore, it is important to know which 
misinformation targets are used, and in which tweets and by which authors, such that people can be inoculated against COVID-19 vaccine misinformation before they are exposed to it. Moreover, because Twitter is the social media platform where most of the conversations about COVID-19 vaccines take place, it is essential to discover automatically tweets that spread misinformation, such that vaccine hesitancy interventions can be delivered to those participating in misinformed conversations. In this paper we present {\sc CoVaxLies}, a corpus of 7,246 tweets judged by language experts to refer to 17 different targets of misinformation about COVID-19 vaccines. The {\sc CoVaxLies} dataset was created using a methodology similar to the one used in the generation of the {\sc COVIDLies}  \cite{covidlies} dataset of tweets, which annotated misinformation about COVID-19. Therefore, they can be used together for learning to identify misinformation about COVID-19 and COVID-19 vaccines. Both {\sc COVIDLies} and
{\sc CoVaxLies} are evolving, as additional targets of misinformation are added and tweets relevant to them are retrieved and judged by experts.
This paper has also explored the need to retrieve tweets relevant to misinformation targets using a combination of retrieval systems, concluding that in this way a larger set of truly relevant tweets are discovered, and can be included in these datasets. 

In this paper, {\sc CoVaxLies} was used to train and evaluate a novel, simple and elegant method for discovering misinformation on Twitter, relying on graph link prediction. This method is enabled by (a) the organization of a Misinformation Knowledge Graph and (b) the availability of link scoring functions from several knowledge embedding models. Our experiments have shown that superior results can be obtained when discovering misinformation using graph link prediction, as compared with neural classification-based methods, yielding an increase of up to 
10\% F1-score. 
The method presented in this paper does not consider
the conversation threads, as many recent misinformation detection methods do, e.g. \cite{gcan2020}, \cite{Ma-GAN}. But {\sc CoVaxLies}  will be extended to account for entire conversation threads on Twitter, allowing us to extend the methodology of misinformation detection presented in this paper, and to
evaluate the impact conversations have on the quality of misinformation detection, when misinformation targets are also considered, an important
aspect that is currently ignored. 

Our future work will also consider the discovery of adoption or rejection of misinformation about COVID-19 vaccines. This will be achieved by relying on the automatic inference of the stance of tweets relative to the misinformation targets. This will allow us to expand our work reported in \cite{our-stance}, where we developed a neural architecture that combined the role of semantic, lexical and affect characteristics of language with the taxonomy of concerns raised by misinformation targets. Along with automatically detecting misinformation about COVID-19 vaccines, the
recognition of the adoption or rejection of that misinformation 
will be a stepping stone in the direction
of developing misinformation inoculation interventions on
social media platforms in the era of COVID-19.

\section*{Conflicts of interest}
There are no conflicts of interest.

\section*{Acknowledgments}


\bibliographystyle{elsarticle-num}
\bibliography{jbi_misinfo}

\end{document}